\documentclass[lettersize,journal]{IEEEtran}
\usepackage{amsmath,amsfonts}
\usepackage{algorithmic}
\usepackage{algorithm}
\usepackage{array}
\usepackage[caption=false,font=normalsize,labelfont=sf,textfont=sf]{subfig}
\usepackage{textcomp}
\usepackage{stfloats}
\usepackage{url}
\usepackage{verbatim}
\usepackage{graphicx}
\usepackage{cite}
\usepackage{enumerate}
\usepackage[backref]{hyperref}
\usepackage{multirow}
\usepackage{multicol}
\usepackage[table,xcdraw]{xcolor}
\usepackage{booktabs}
\usepackage{makecell}
\usepackage[flushleft]{threeparttable}
\usepackage[normalem]{ulem}
\useunder{\uline}{\ul}{}

\begin{document}

\title{
Normal Reference Attention and Defective Feature Perception Network for Surface Defect Detection}
\author{Wei Luo,~\IEEEmembership{Student Member,~IEEE},~Haiming Yao,~\IEEEmembership{Student Member,~IEEE},~Wenyong Yu,~\IEEEmembership{Member,~IEEE}
\thanks{Manuscript received XX XX, 20XX; revised XX XX, 20XX. This study was financially supported by the National Natural Science Foundation of China (Grant No. 51775214) (Corresponding author: Wenyong Yu.)}
\thanks{Wei Luo and Wenyong Yu are with the State Key Laboratory of Digital Manufacturing Equipment and Technology, School of Mechanical Science and Engineering, Huazhong University of Science and Technology, Wuhan 430074, China (e-mail: u201910709@hust.edu.cn, ywy@hust.edu.cn).}
\thanks{Haiming Yao is with the State Key Laboratory of Precision Measurement Technology and Instruments, Department of Precision Instrument, Tsinghua University, Beijing 100084, China (e-mail: yhm22@mails.tsinghua.edu.cn).}}

\maketitle

\markboth{Journal of \LaTeX\ Class Files,~Vol.~14, No.~8, August~2021}%
{Shell \MakeLowercase{\textit{et al.}}: A Sample Article Using IEEEtran.cls for IEEE Journals}

\begin{abstract}

Visual anomaly detection plays a significant role in the development of industrial automatic product quality inspection. As a result of the utmost imbalance in the amount of normal and abnormal data, growing attention has been given to unsupervised methods for defect detection. Although existing reconstruction-based methods have been widely studied recently, establishing a robust reconstruction model for various textured surface defect detection remains a challenging task due to homogeneous and nonregular surface textures. In this paper, we propose a novel unsupervised reconstruction-based method called the normal reference attention and defective feature perception network (NDP-Net) to accurately inspect a variety of textured defects. Unlike most reconstruction-based methods, our NDP-Net first employs an encoding module that extracts multi scale discriminative features of the surface textures, which is augmented with the defect discriminative ability by the proposed artificial defects and the novel pixel-level defect perception loss. Subsequently, a novel reference-based attention module (RBAM) is proposed to leverage the normal features of the fixed reference image to repair the defective features and restrain the reconstruction of the defects. Next, the repaired features are fed into a decoding module to reconstruct the normal textured background. Finally, the novel multi scale defect segmentation module (MSDSM) is introduced for precise defect detection and segmentation. In addition, a two-stage training strategy is utilized to enhance the inspection performance. The comprehensive experimental results indicate that the NDP-Net method achieves state-of-the-art inspection performance with an area under the receiver operating characteristic curve (ROCAUC) of 98.81\% for defect detection and 98.54\% for defect segmentation over all 5 textured surfaces on the MVTec AD dataset and 99.59\% on 10 challenging texture surfaces on the DAGM dataset. Furthermore, NDP-Net also achieves satisfactory results in practical industrial applications.
\end{abstract}


\section{Introduction}
\IEEEPARstart{I}{n } manufacturing industries, surface defects are common in substantial industrial products, such as steel \cite{steel}, wood \cite{wood}, and TFT-LCDs \cite{TFT-LCD}, because of the raw materials employed and complex manufacturing processes. These defects are localized destruction of the texture surface structure or texture surface pattern, which not only directly affects the experiences of the users but also may cause industrial accidents. Therefore, to achieve product quality management and improve the experience of users, surface defect detection has become an indispensable component during the manufacturing process.

In recent years, a variety of defect detection methods have been presented, which can be loosely classified into two main approaches: conventional methods and deep learning-based methods. Most conventional methods \cite{PHOT,LCA,TEXTEMS} utilize handcrafted features to obtain texture defect representations, resulting in a low inspection efficiency, a high false detection rate, and a poor consistency on different texture surfaces. However, as a data-based method, deep learning-based methods have the ability to extract features automatically by training a large amount of textured surfaces, with good robustness.

Deep learning-based methods can be divided into supervised and unsupervised methods, in which whether the training dataset contains labeled defective images is the judgement criterion. Supervised methods have made a good progress in the field of defect detection in recent years. Dong \textit{et al.} \cite{PGANet} proposed a pyramid feature fusion module aiming to obtain fused features at five resolutions. Cheng \textit{et al.} \cite{RetinaNet} proposed a channel attention mechanism with the aim to retain more detail information. Although supervised methods have performed well with regards to surface defect detection, collecting a comprehensive and balanced set of defective images is still the main difficulty in regard to the surface defect detection problem because the number of normal images is much larger than that of the defective images in the actual industrial field.

In contrast, unsupervised methods are expected to solve this problem because they only require unlabeled defect-free samples for training. In recent years, embedding-based \cite{PatchSVDD,SPADE,GCPF} and reconstruction-based \cite{AnoGan,MSCDAE,MSFCAE} methods have been investigated for unsupervised defect detection and segmentation. Embedding-based methods detect anomalies by mapping the image to the feature domain. The Patch-SVDD framework \cite{PatchSVDD} builds the embedded feature domain by leveraging self-supervised learning. The SPADE \cite{SPADE} method presents adequate performance on defect detection. The GCPF \cite{GCPF} method leverages multiple independent Gaussian clustering to inspect anomalies. These methods can perform image-level defect detection but not accurate pixel-level defect segmentation. Reconstruction-based methods leverage the deep convolutional neural network to reconstruct the anomaly image to a normal texture background by restoring the defects. Consequently, the defective regions are obtained by the difference between the original input and the reconstructed input. The AnoGAN \cite{AnoGan} leverages the generative adversarial network to learn the distribution of defect-free samples, thus generating the normal texture background. Mei \textit{et al}.\cite{MSCDAE} proposed a model that is robust to noise, called multiscale convolutional denoising autoencoder (MSCDAE), to inspect texture defects. Yang \textit{et al}. \cite{MSFCAE} proposed a multiscale feature-clustering-based fully convolutional autoencoder (MSFCAE) method for various types of texture defect detection. These methods cannot convert the defective feature to the normal feature, which leads to the reconstruction of defects. FMR-Net \cite{FMR-Net} proposes a memory feature module and a global feature rearrangement module to restore the defects. However, a key point ignored by FMR-Net is the feature maps generated by memory feature module may also exist defects. Then, the FMR-Net model reconstructs texture images at patch level, which leads to its slow inference speed. In addition, the FMR-Net method does not have pixel-level defect perception capabilities, resulting in mis-inspections.

As discussed above, we can conclude that the two key issues of the reconstruction-based methods are repairing the defects to the normal textures and accurately segmenting defects based on the input images and the corresponding reconstructed images. To address the first issue, instead of leveraging feature map generated by memory feature module in FMR-Net \cite{FMR-Net} to restore defects, the novel reference-based attention module (RBAM) is proposed to utilize the features of the fixed reference normal sample to repair the defective features. To tackle the second issue, a novel pixel-level defect perception loss and multiscale defect segmentation module (MSDSM) are proposed, where the former is leveraged to augment the encoder with a defect discriminative ability and the latter is utilized for accurate defect segmentation on the basis of multi-scale features of the input images and the corresponding reconstructed images. Therefore, the entire NDP-Net framework can repair and accurately segment various textured defects simultaneously.

The main contributions of our work are as follows.
\begin{itemize}
    \item We propose a novel pixel-level defect perception loss function, which enables the model to obtain the pixel-level defect perception capability.
    \item We propose a novel reference-based attention module (RBAM), which not only repairs the defective features but also suppresses the residual of the defective features caused by skip-connections.
    \item We propose a novel multi-scale defect segmentation module (MSDSM) to segment defects more accurately.
    \item The proposed method achieves an improved performance on the defect detection results on the MVTec AD dataset \cite{MVTEC} compared with the outstanding methods, with an area under the receiver operating characteristic curve (ROCAUC) of 98.81\% for defect detection and 98.54\% for defect segmentation over all 5 textured surface and 99.59\% on 10 challenging texture surfaces on the DAGM dataset \cite{DAGM}. Furthermore, NDP-Net also achieves satisfactory results in practical industrial applications.
\end{itemize}

The remainder of this article is organized as follows. The related work about unsupervised methods for defect detection and segmentation is stated in Section \MakeUppercase{\romannumeral2}. The proposed NDP-Net is introduced in detail in Section \MakeUppercase{\romannumeral3}. In Section \MakeUppercase{\romannumeral4}, comprehensive experiments are conducted to analyze the performance of NDP-Net. Finally, the conclusion and future work are discussed in Section \MakeUppercase{\romannumeral5}.

\begin{figure*}[!t]
\centering
\includegraphics[scale=1]{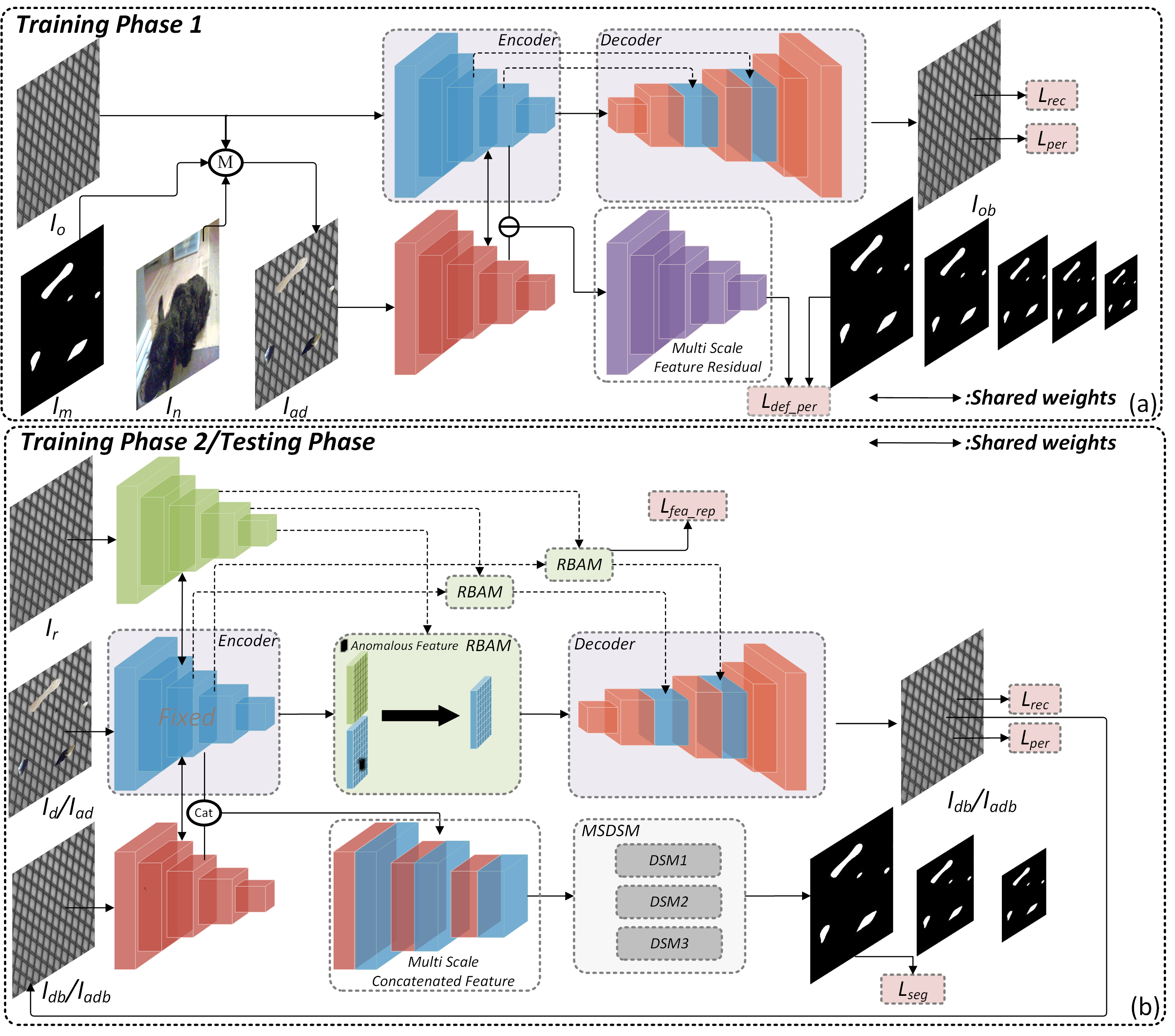}
\caption{Overall architecture of the proposed NDP-Net methods. NDP-Net consists of an encoder, an RBAM, a decoder and an MSDSM. (a) During Training Phase 1, defect-free images and artificial defect images are propagated forward. Artificial defect images and the pixel-level defect perception loss ($L_{def\_per}$) are adopted to make the encoder have a pixel\_level normal/abnormal discrimination ability. (b) During Training Phase 2, the fixed reference defect-free image and artificial defect images are propagated forward, and the weights of the encoder are fixed. The RBAM utilizes the features of the fixed reference image to repair the features of artificial defect images, and then the repaired features are inputted or concatenated to the corresponding layers in the decoder. The MSDSM leverages the concatenated features of the original artificial defect images and the corresponding reconstructed images for defect segmentation. The pink boxes indicate losses.}
\label{fig_1}
\end{figure*}

\section{Relate Works}
In this section, we introduce related works regarding the unsupervised methods for the defect detection and segmentation. The unsupervised methods used for the defect detection and segmentation can be mainly divided into two categories: embedding-based and reconstruction-based methods.
\subsection{Embedding-based Methods}

The key to the embedding-based methods is to properly construct a discriminative embedding feature domain, where the defective samples are embedded far away from the normal centers. The CNN\_Dict method \cite{CNN_Dict} leverages the PCA \cite{PCA} to reduce the dimensionality of the patch-level feature extracted by ResNet18 \cite{ResNet18}, and then utilizes k-means \cite{kmeans} to cluster the compressed features into different representative groups. The distances of the testing features to the clustering centers are computed as the anomaly scores. Bergmann \textit{et al}. \cite{ST} proposed a Student-Teacher framework to inspect defects by the difference between the output of student network and teacher network and the regression errors. The Patch SVDD \cite{PatchSVDD} method is proposed to locate defects by leveraging self-supervised learning, but the extraction and retrieval of the patch-level feature are time-consuming. The SPADE \cite{SPADE} method utilizes the K-NN \cite{KNN} method to retrieve similar features from the pre-stored normal features in the testing phase, which shows an effective performance on anomaly detections. Similar to the Patch SVDD method, the two retrievals required by the SPADE method consume much more computation time and online memory storage. The GCPF \cite{GCPF} method is proposed to inspect defects using multiple independent Gaussian clustering, which effectively reduces the computation time and online memory storage requirement. The above embedding-based methods either require much more computation time and memory storage or cannot segment anomalies accurately.
\subsection{Reconstruction-based Methods}

Reconstruction-based models have been extensively investigated for defect detection and segmentation. Reconstruction-based methods utilize only defect-free images for training and are expected to restore the defects during the testing phase. Then, the differences between the original images and their reconstructed images are computed to locate the defects. Auto Encoder (AE) \cite{AE} is a typical reconstruction-based model. Images reconstructed by AE are almost blurred and lack structural information. Therefore, the structural similarity \cite{SSIM} loss function is applied to enhance the common AE. Meanwhile, the common AE has a strong generalization capability, resulting in the reconstruction of defects. The MemAE \cite{MemAE} method proposes a memory module that mitigates the drawback of over-generation. Luo \textit{et al}. \cite{CMA-AE} proposed a clear memory-augmented autoencoder (CMA-AE) to make the model have both a strong normal background reconstruction and abnormal foreground restoration ability. Recently, perceptual distance \cite{perceptual_distance} demonstrated that deep features can measure the similarity between two images in a way that coincides with human judgment. The TrustMAE \cite{TrustMAE} method
combines the MemAE and the perceptual distance for defect segmentation. The RIAD \cite{RIAD} method randomly removes partial patches of different sizes and reconstructs the missing information from partial inpainting. Recently, GANs \cite{GANs} have shown outstanding generative ability. The AnoGAN \cite{AnoGan} leverages the GAN generator to inspect defects. However, the AnoGAN and original GANs lack mapping from the image domain to the feature domain. To address this problem, many GAN-based methods have been proposed. The GANomaly \cite{Ganomaly} method proposes an encoder-decoder-encoder framework that optimizes the model in both the image space and feature space. To achieve a more detailed reconstruction and speed up the convergence of the model, the Skip-GANomaly \cite{skip-ganomaly} combines the GANomaly and the skip-connection. However, the above methods cannot convert defective features to normal features. The ST-MAE \cite{STMAE} repairs defective features from the perspective of a deep feature transition. AFEAN \cite{AFEAN} proposes a global context feature editing module to eliminate the effect of defect reconstructions. The defective feature residual caused by skip-connection still exists in AFEAN and Skip-GANomaly. To address this problem, the FMR-Net \cite{FMR-Net} utilizes memory-generated features to repair the defective features. Then, the repaired features are fed into the decoder through the skip-connection. However, the key point that the FMR-Net misses is that the memory-generated features may also contain defective features.

Overall, converting the defective features to normal features is still a challenge for reconstruction-based methods. To address this challenge, the NDP-Net framework is proposed in this article. The NDP-Net learns to repair defective features through the novel RBAM. Thus, NDP-Net can reconstruct the defective images to normal images, thereby achieving a state-of-the-art performance in defect detection and segmentation.

\section{Proposed NDP-Net Framework}
In this section, the proposed NDP-Net framework is introduced in detail. First, the overall architecture of NDP-Net is briefly introduced. Then, its main modules, including the encoder with a pixel-level defect perception capability, the reference-based attention module (RBAM), and the multi-scale defect segmentation module (MSDSM), are presented in detail. Finally, the training and testing procedures of the NDP-Net are introduced.

\begin{figure}[!t]
\centering
\includegraphics[scale=1]{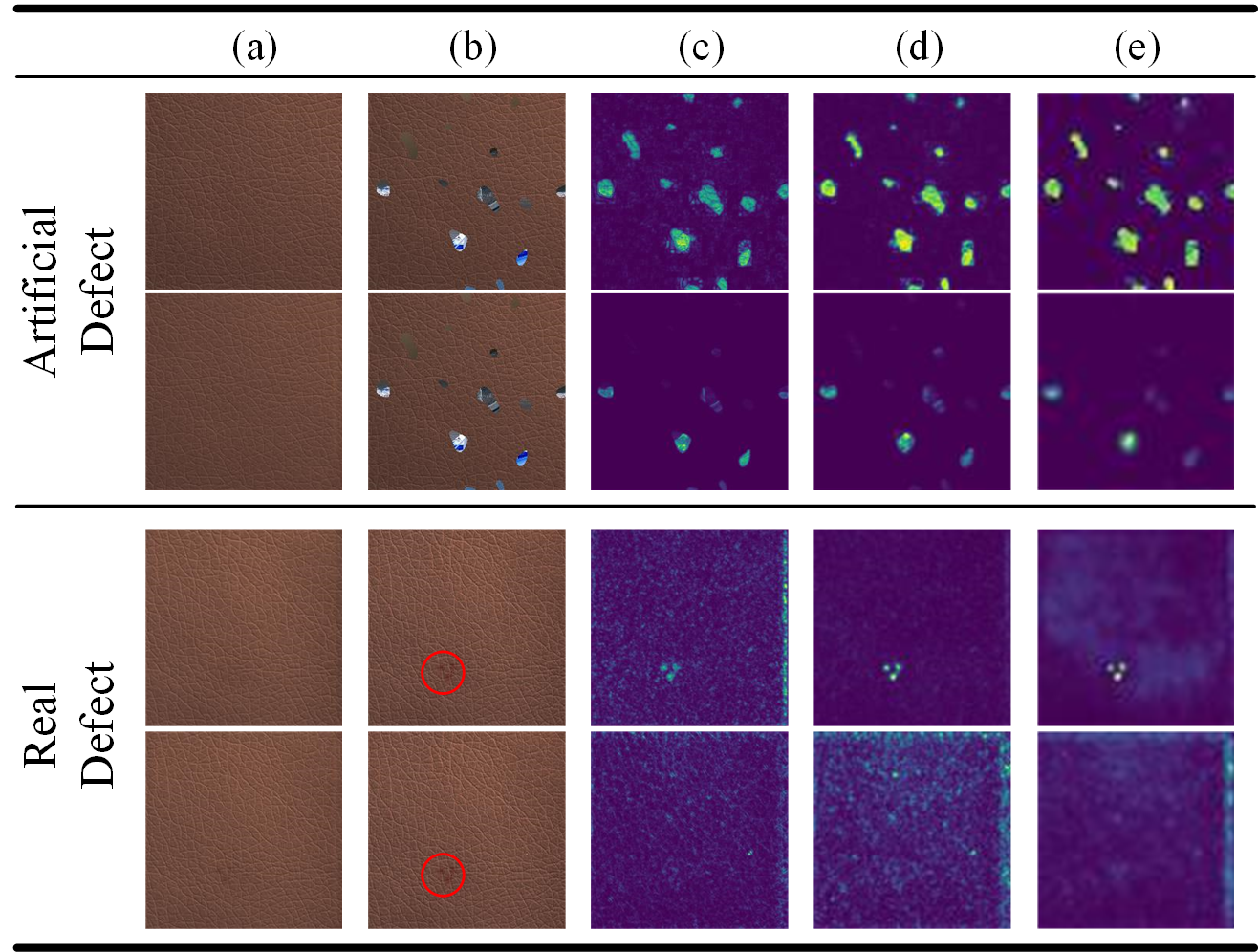}
\caption{Examples of the effect of the pixel-level defect perception loss function. The first row is the result of the encoder trained using $L_{def\_per}$, and the second row is the result of the traditional encoder. For artificial defects, (a) defect-free image, (b) artificial defective image. For a real defect, (a) reconstructed image, (b) real defective image. (c), (d), and (e) denote the difference between the feature map of (a) and the feature map of (b) extracted by the 1st, 2nd, and 3rd layers of the encoder, respectively. The above real defective image is from the MVTec AD dataset \cite{MVTEC}.}
\label{fig_2}
\end{figure}

\subsection{Overall Architecture of NDP-Net}

The key point of the reconstruction-based methods is to convert the defective features to normal features. Current reconstruction-based methods
do not have a strong defect repairing capability. Meanwhile, they also lack pixel-level normal/abnormal discriminatory capabilities. In this study, a novel NDP-Net method is proposed to address these challenges.

The overall architecture of NDP-Net is shown in Fig. \ref{fig_1}. RBAE consists of four modules: the encoder (for extracting multi-scale semantic features from images), RBAM (for repairing defects in the feature domain), decoder (for reconstructing the normal texture background from the features), and MSDSM (for defect segmentation). The training procedure is divided into two phases. In the first phase, the artificial defect images and the defect-free images are inputted into the encoder. The pixel-level defect perception loss function ($L_{def\_per}$) is leveraged to make the encoder have a pixel-level normal/abnormal discriminatory ability. Then, only the features of defect-free images are inputted into the decoder to reconstruct the textured background. In the second phase, the weights of the encoder are fixed, and the encoder extracts features of the fixed reference image and artificial defect images. The RBAM utilizes the features of the reference image to repair the features of artificial defect images. Then, the repaired features are inputted into the decoder to reconstruct the textured background. Then, the reconstructed images and the original artificial defect images are reinput into the encoder, and the features of the first three layers are separately concatenated. Finally, MSDSM employs concatenated features to obtain multi-scale defect segmentation maps. The procedure for the testing phase is consistent with the second training phase.

\subsection{Encoder with a Pixel-Level Defect Perception Capability}

The traditional encoder is used to compress the image information, and it is trained with only defect-free images. Consequently, this encoder is not capable of perceiving defects at the pixel level. To address this challenge, we adopt pairs of images: a defect-free image $I_{o}$ and its artificial defective image $I_{ad}$. The artificial defects are generated in a similar way to DRAEM \cite{draem}. As shown in Fig. \ref{fig_1}(a), the artificial defective images are generated by combining the defect-free images $I_{o}$, the natural images $I_{n}$ from the ImageNet dataset \cite{ImageNet} and random masks $I_{m}$:
\begin{equation}
    I_{ad} = I_{o}\odot(1 - {I_m}) + {I_{n}}\odot{I_{m}}
    \label{eq1}
\end{equation}
where $\odot$ denotes the dot product operation, ${I_o},{I_m},{I_n},{I_{ad}} \in {R^{H \times W \times C}}$ and
H, W, and C represent the height, width, and number of the channels, respectively. Moreover, as shown in Fig. \ref{fig_1}, we propose a novel pixel-level defect perception loss function to make the encoder equipped with the ability discriminate normal/abnormal regions at the pixel level. We let ${\Phi _l} \in {R^{{H_l} \times {W_l} \times {C_l}}}$ represent the features extracted from the $l$th layer of the encoder. The pixel-level defect perception loss function is defined as:
\begin{equation}
{L_{def\_per}} = \frac{1}{L}\sum\limits_{l = 1}^L {{{\left\| {N({\rm M}({{\left\| {{\Phi _l}({I_o}) - {\Phi _l}({I_{ad}})} \right\|}^2})) - I_m^l} \right\|}^2}} 
\label{eq2}
\end{equation}
where $I_{m}^{l}$ represents the random mask resized to $(H_{l}, W_{l})$, L represents the number of layers in the encoder, $M(\cdot)$ represents the mean operation at the channel dimension, and $N(\cdot)$ represents the normalization operation. The effect of $L_{def\_per}$ is shown in Fig. \ref{fig_2}. As we can see, $L_{def\_per}$ indeed equips the encoder with the capability of pixel-level defect perception, and this capability can be generalized to real defects.

\begin{figure*}[!t]
\centering
\includegraphics[scale=0.92]{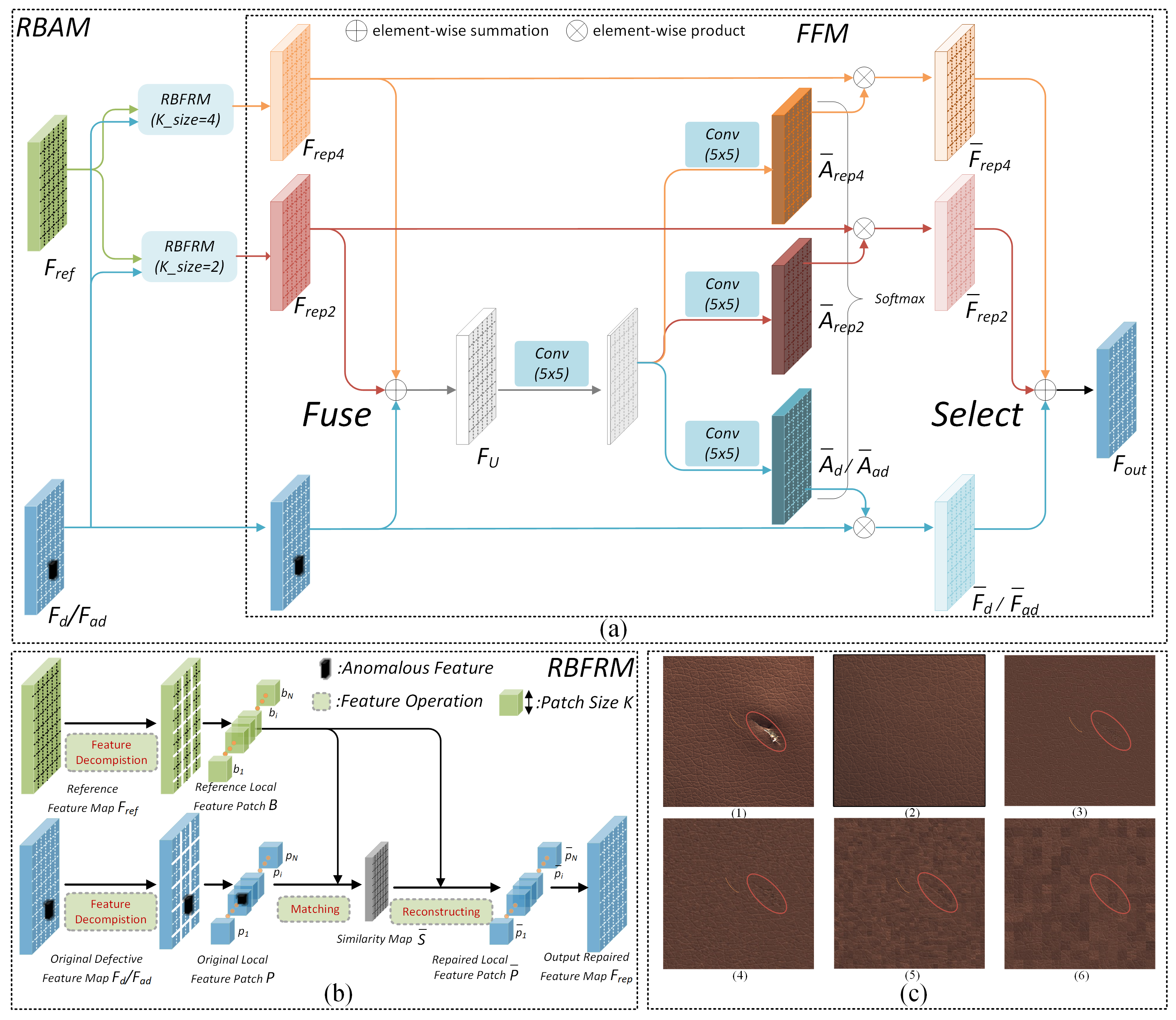}
\caption{Illustration of the RBAM. (a) Architecture of the RBAM. The RBAM consists of two modules: RBFRM and FFM. (b) Illustration of the RBFRM. The RBFRM utilizes the reference features to repair the defective features. (c) Effect of the patch size $K$ on RBFRM at the image level. (1) Defective image. (2) Fixed reference defect-free image. (3), (4), (5), and (6) denote the repaired image using RBFRM whose $K$ is 2, 4, 8, and 16, respectively. The orange dotted lines indicate the textures. The red ellipses indicate the defects.}
\label{fig_3}
\end{figure*}

\subsection{Reference-Based Attention Module}
The mainstream reconstruction-based methods cannot address the defective features, resulting in the reconstruction of defects.
Moreover, some methods adopt the skip-connection operation to
pursue more detailed information, but this brings defective features from the encoder into the decoder. Inspired by the contextual attention \cite{CA}, we proposed RBAM to convert the defective features to normal features and suppress the defective feature residuals in skip-connections to address the above problems. 
As shown in Fig. \ref{fig_3} (a), the RBAM consists of the reference-based feature repair module (RBFRM) and the feature fusion module (FFM). The RBFRM and FFM are described in detail as follows.

\subsubsection{Reference-Based Feature Repair Module}
Addressing the defective features is a major challenge in defect detection. The FMR-Net \cite{FMR-Net} method leverages memory-generated features to repair the defective features. However, the feature maps generated by the memory mechanisms \cite{MemAE} may also contain defective features. Therefore, we choose a fixed reference defect-free image to repair the defects. As shown in Fig. \ref{fig_3} (b), we name the feature maps of the fixed reference defect-free image extracted by the encoder the reference feature map and denote them as ${F_{ref}} \in {R^{{H_F} \times {W_F} \times {C_F}}}$, where $H_F$, $W_F$ and $C_F$ represent the height, width and number of channels, respectively, of the feature maps. Then, we divide the reference feature map into non-overlapping patches, which are named reference local feature patches and are denoted by $B = \{ {b_1},{b_2}, \cdots ,{b_N}\}$, where ${b_i} \in {R^{K \times K \times {C_F}}}$, $i \in (1, \cdots ,\frac{{{H_F}}}{K} \times \frac{{{W_F}}}{K})$ and $K$ is the patch size. Similarly, we denote the original real/artificial defective feature map by $F_d/F_{ad}$ and apply the same feature decomposition operation described above for a set of original local feature patches $P = \{ {p_1},{p_2}, \cdots ,{p_N}\}$, where the size of $p_i$ is the same as $b_i$. To match the defective patches with reference defect-free patches, we calculate the cosine similarity between them as:
\begin{equation}
{S_{i,j}} = \left\langle {\frac{{{p_i}}}{{\left\| {{p_i}} \right\|}},\frac{{{b_j}}}{{\left\| {{b_j}} \right\|}}} \right\rangle 
\label{eq3}
\end{equation}
where $\left\|  \cdot  \right\|$ represents the modulus of a tensor, and $S_{i,j}$ represents the similarity map $S$ at the location $(i, j)$, which calculates the similarity of the $ith$ original defective feature patch and the $jth$ reference feature patch. Then, we apply the softmax function to obtain the normalized similarity:
\begin{equation}
    {{\bar S}_{i,j}} = \frac{{\exp ({S_{i,j}})}}{{\sum\limits_{j = 1}^N {\exp ({S_{i,j}})} }}
    \label{eq4}
\end{equation}

Finally, we utilize the reference local feature patches $B$ and the normalized similarity map ${\bar S}$ to reconstruct the output repaired local feature patches $\bar P = \{ {{\bar p}_1},{{\bar p}_2}, \cdots ,{{\bar p}_N}\} $, described as follows:
\begin{equation}
{{\bar p}_i} = \sum\limits_{j = 1}^N {{b_j} \cdot {{\bar S}_{i,j}}} 
\label{eq5}
\end{equation}

Then, we obtain the output repaired feature map $F_{rep}$ by directly composing all the repaired local feature patches.

\subsubsection{Feature Fusion Module}
The hyperparameter patch size $K$ of the RBFRM influences the quality of feature repairs. As shown in Fig. \ref{fig_3} (c), the larger the patch size $K$ is, the better the defective features are suppressed. However, as $K$ increases, the repaired features retain less of the original textured features and gradually exhibit a "checkerboard" effect. To address this problem, inspired by SKNet \cite{SKNet}, we propose the feature fusion module (FFM).

The architecture of the FFM is shown in Fig. \ref{fig_3} (a). We denote the feature map repaired by RBFRM whose K is set to 4 as $F_{rep4}$, and $F_{rep2}$ is defined similarly. First, we fuse the information of $F_{d}$, $F_{rep2}$ and $F_{rep4}$ via an elementwise summation:
\begin{equation}
{F_U} = {F_{ad}} + {F_{rep2}} + {F_{rep4}}
\label{eq6}
\end{equation}

Then, we compress the channel dimension to 64 by using a convolution operation with a $m$$\times$$m$ kernel, whose aim is to obtain more important channel information. Next, the three different $m$$\times$$m$ convolution kernels are applied to obtain attention maps $A_{rep4}$, $A_{rep2}$ and $A_{d}/A_{ad}$. We assume that the size of $m$ should be larger than the maximum $K$. In this paper, the maximum $K$ is set to 4, as mentioned above; thus, $m$ is set to 5. Specifically, we apply a softmax function on the channel-wise digits for the normalized attention maps:
\begin{equation}
\begin{split}
\bar A_{rep4}^{i,j,k} = \frac{{\exp (A_{rep4}^{i,j,k})}}{{\exp (A_{rep4}^{i,j,k}) + \exp (A_{rep2}^{i,j,k}) + \exp (A_{ad}^{i,j,k})}},\\
\bar A_{rep2}^{i,j,k} = \frac{{\exp (A_{rep2}^{i,j,k})}}{{\exp (A_{rep4}^{i,j,k}) + \exp (A_{rep2}^{i,j,k}) + \exp (A_{ad}^{i,j,k})}},\\
\bar A_{ad}^{i,j,k} = \frac{{\exp (A_{ad}^{i,j,k})}}{{\exp (A_{rep4}^{i,j,k}) + \exp (A_{rep2}^{i,j,k}) + \exp (A_{ad}^{i,j,k})}}
\end{split}
\label{eq7}
\end{equation}
where $\bar A_{rep4}^{i, j, k}$, $\bar A_{rep2}^{i, j, k}$ and $\bar A_{ad}^{i, j, k}$ represent the $i$th row, $j$th column, $k$th element of $\bar A_{rep4}$, $\bar A_{rep2}$ and $\bar A_{ad}$, respectively. It is obvious that $\bar A_{rep4 }+ \bar A_{rep2} + \bar A_{ad} = 1$. The three normalized attention maps are employed to select the information of $F_{rep4}$, $F_{rep2}$, and $F_{ad}$. Finally, the output feature map $F_{out}$ can be calculated as follows:
\begin{equation}
{F_o} = {{\bar A}_{rep4}} \odot {F_{rep4}} + {{\bar A}_{rep2}} \odot {F_{rep2}} + {{\bar A}_{ad}} \odot {F_{ad}}
\label{eq8}
\end{equation}
where $\odot$ denotes the dot product operation.
\begin{figure}[!t]
    \centering
    \includegraphics[scale=1]{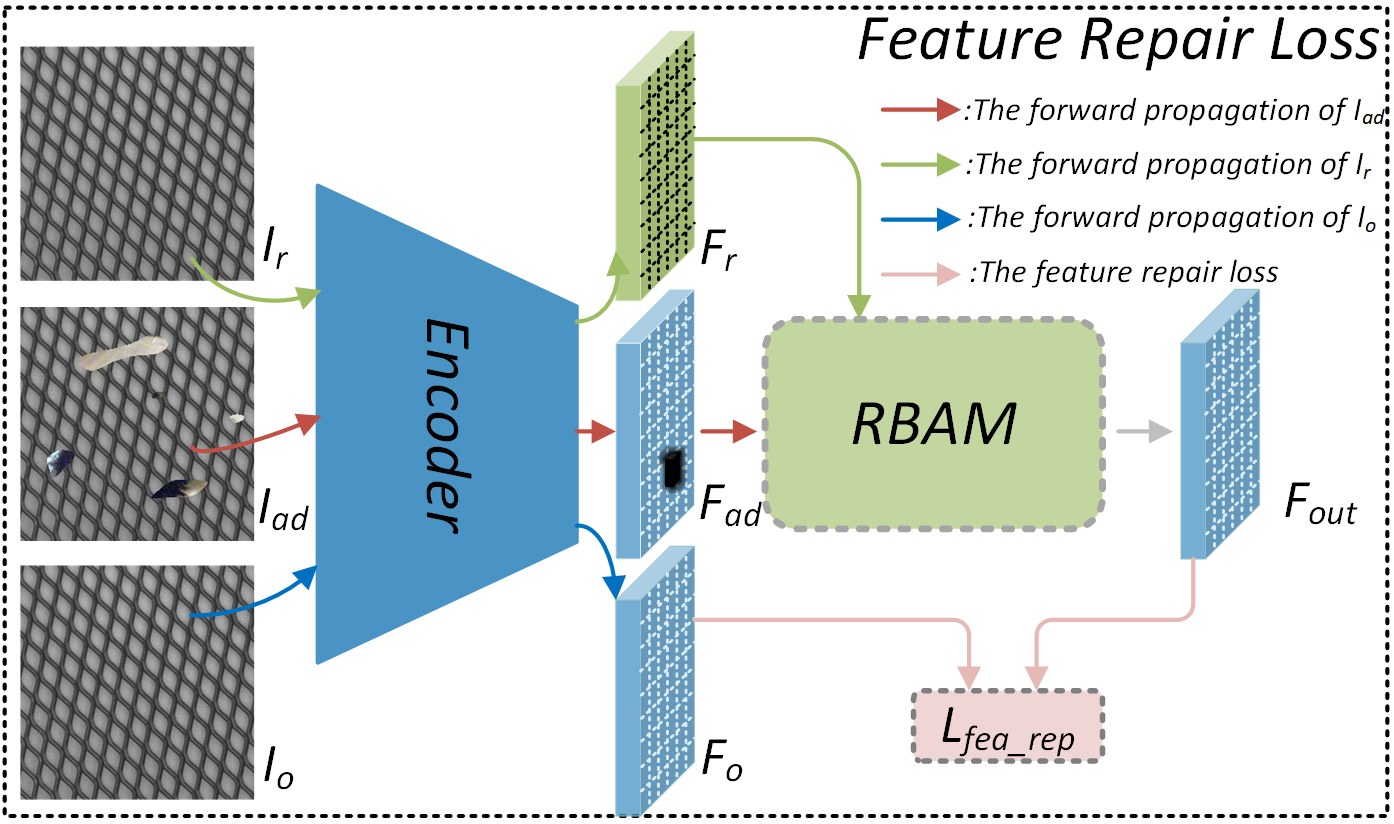}
    \caption{Schematic of the feature repair loss function. The feature repair loss function is utilized to guide the RBAM to repair the defective features.}
    \label{fig_4}
\end{figure}

\subsubsection{Feature Repair Loss Function}
To guide the RBAM to convert the defective features to the normal features, the feature repair loss function $L_{fea\_rep}$ is proposed. As shown in Fig. \ref{fig_4}, we can obtain the feature map of the original defect-free image, which is denoted as $F_{o}$. There is no doubt that we hope that the artificial defective feature map $F_{ad}$ can be repaired to the corresponding original defect-free feature map $F_{o}$. Therefore, we simply calculate the mean squared error between $F_{o}$ and $F_{ad}$ as the feature repair loss function:
\begin{equation}
{L_{fea\_rep}} = \frac{1}{{{l_2} - {l_1}}}\sum\limits_{l = {l_1}}^{{l_2}} {{{\left\| {F_{out}^l - F_o^l} \right\|}^2}} 
\label{eq9}
\end{equation}
where $F_{o}^l$ denotes $F_{o}$ extracted by the $l$th layer of the encoder, likewise $F_{out}^l$, and $l_{1}$ and $l_{2}$ represent the first and last layer where the RBAM module is embedded into the encoder, respectively. Next, we discuss the values of $l_{1}$ and $l_{2}$.

\subsubsection{Discussion about How the RBAMs are Embedded onto the Different Scales of NDP-Net}
There are five layers of the encoder. With the deeper encoder layer, the receptive fields of the feature maps become larger, and the feature maps become more semantic but less detailed. There is no doubt that local feature patches require high-level semantic representation. In addition, more texture detail information can be useful for the network to accurately reconstruct the background. Therefore, we need to balance the semantic representation and texture detail information. In this paper, we employ the RBAMs in the three deepest layers of NDP-Net, which can obtain a good balance between the semantic representation and the texture detail information. Therefore, $l_1$ and $l_2$ are set to 3 and 5, respectively.

\begin{figure}[!t]
    \centering
    \includegraphics[scale=1]{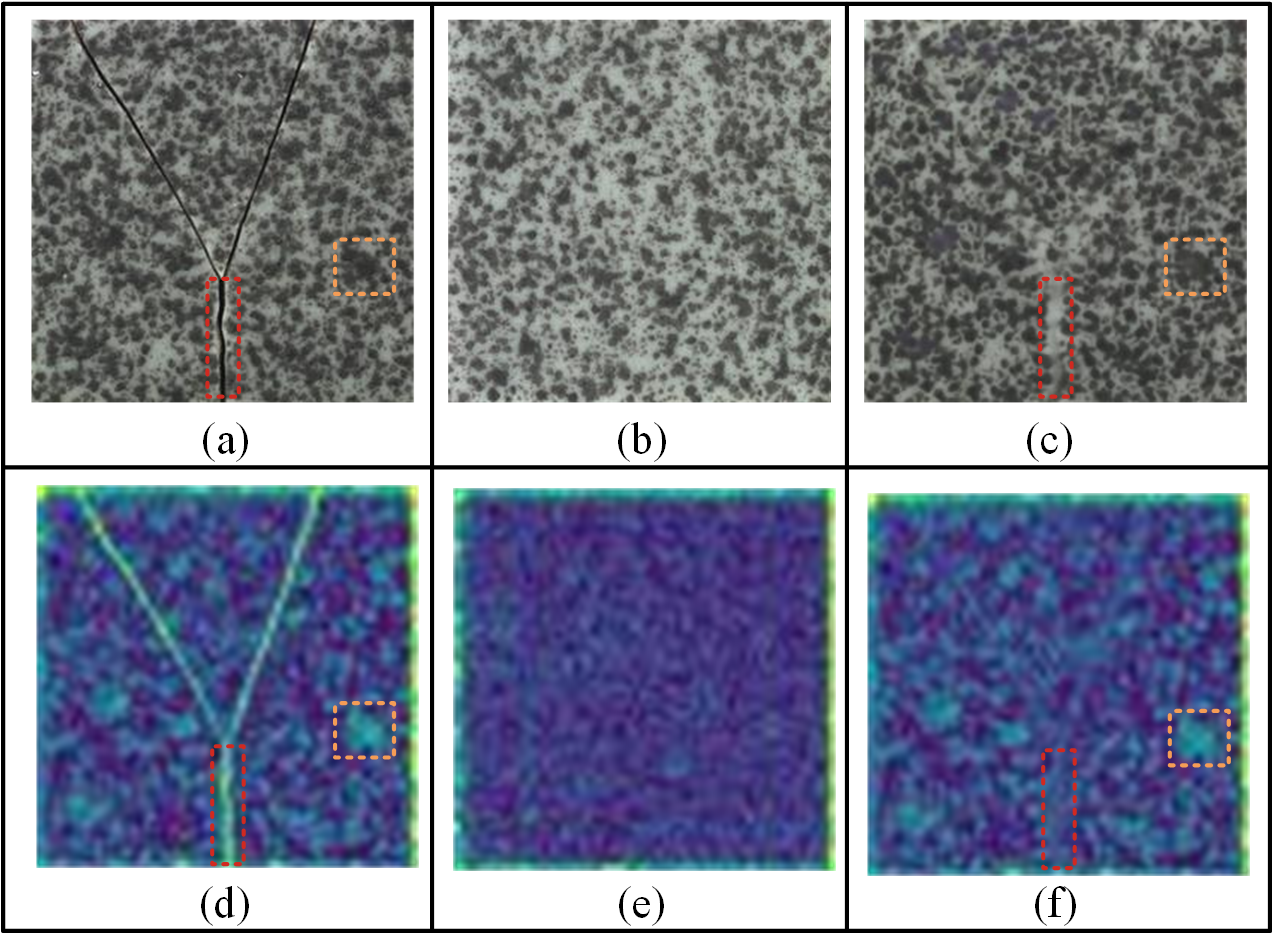}
    \caption{Effect of the RBAM. The orange-dotted boxes indicate the textures. The red-dotted boxes indicate the defects. (a) The defective image. (b) The fixed reference defect-free image. (c) The reconstructed image. (d) The feature map of the defective image extracted by the $3$rd layer of the encoder. (e) The feature map of the fixed reference defect-free image extracted by the $3$rd layer of the encoder. (f) The output repaired the feature map. The above defective image is from the MVTec AD dataset \cite{MVTEC}.}
    \label{fig_5}
\end{figure}

Finally, as shown in Fig. \ref{fig_5}, the RBAM can not only suppress the defective features but also retain the normal texture features, which is useful to achieve a more accurate texture background reconstruction.

\subsection{Multi-Scale Defect Segmentation Module}
For the defect detection task, there are two important issues: one is converting the defective features to normal textured features and thus suppressing the reconstruction of defects, and the other is accurately inspecting and segmenting defects through the original defective images and the corresponding reconstructed images. The former can be solved by the RBAM proposed above, while the latter is still a challenge. The pixel difference between the original image and its reconstructed image is commonly used for defect detection and segmentation, but there are two critical problems in this method. The first is that individual pixels have no semantic information. However, discriminating normal/abnormal pixels is based on rich contextual information. The second is that the size of industrial defects, in reality, is multi-scale; thus, achieving accurate defect detection and segmentation requires multi-scale information. To solve the two problems in the pixel difference method, the multi-scale defect segmentation module (MSDSM) is proposed.

\begin{figure}[!t]
    \centering
    \includegraphics[scale=0.95]{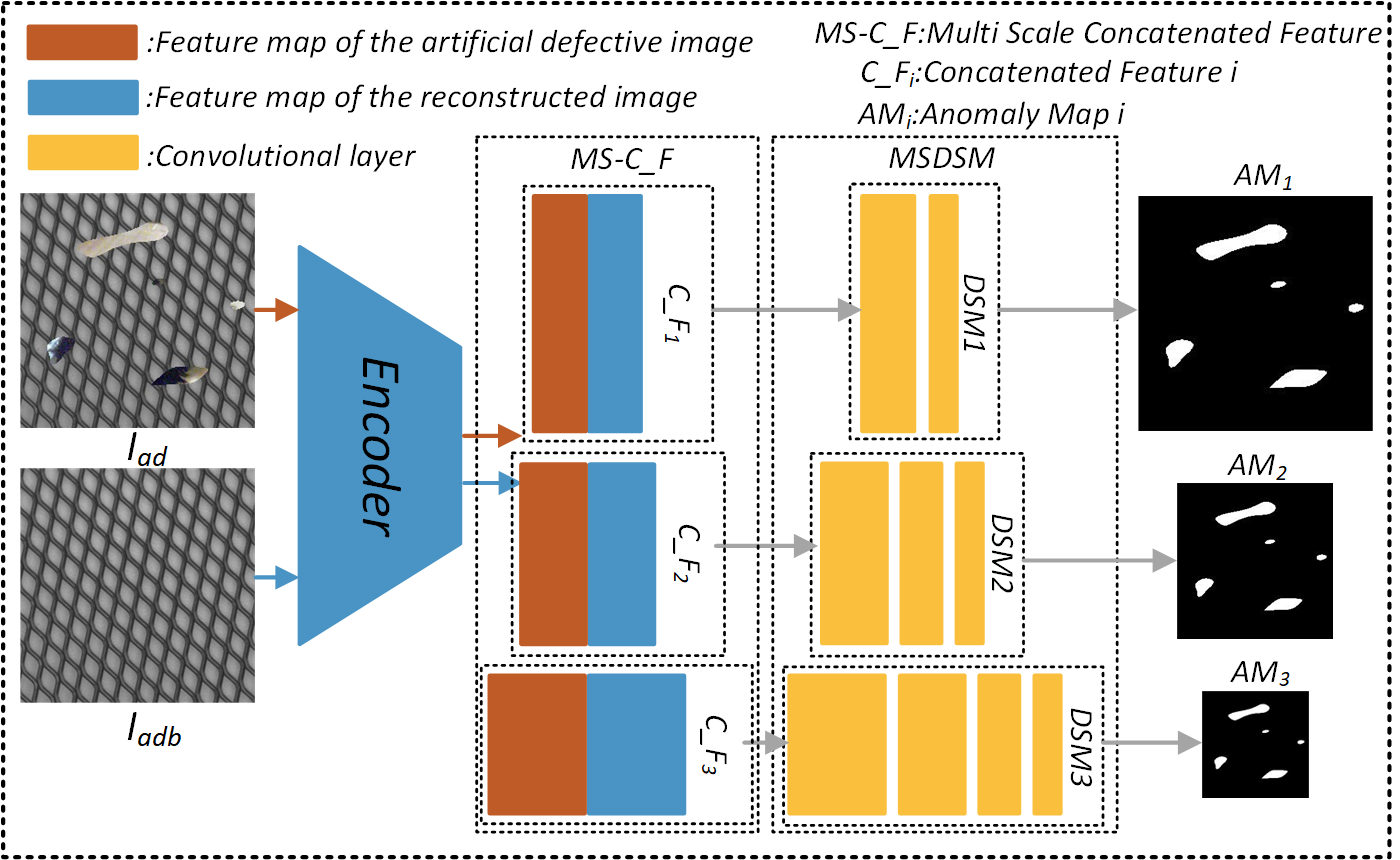}
    \caption{Schematic of the Multi-Scale Defect Segmentation Module (MSDSM).}
    \label{fig_6}
\end{figure}

As shown in Fig. \ref{fig_6}, the MSDSM has three sub-networks, namely, DSM1, DSM2, and DSM3, which are used to detect defective features at different scales. First, the artificial/real defective image ($I_{ad}/I_{d}$) and the corresponding reconstructed image ($I_{adb}/I_{db}$) are re-inputted into the encoder to obtain the multi-scale concatenated features $MS - {C_\_}F = \left\{ {{C_\_}{F_1},{C_\_}{F_2},{C_\_}{F_3}} \right\}$, where ${C_\_}F_1$, ${C_\_F}_2$ and ${C_\_}F_3$ are the corresponding concatenated features extracted by the first three layers of the encoder. Then, the MSDSM utilizes $MS-C_\_F$ to obtain the anomaly maps $AM=\left\{AM_1, AM_2, AM_3\right\}$ at different scales. In addition, to increase the robustness of the MSDSM, we apply the focal loss \cite{focalloss} as the segmentation loss:
\begin{equation}
{L_{seg}} = \frac{1}{3}\sum\limits_{i = 1}^3 {Los{s_{focal}}(I_m^i,A{M_i})} 
\label{eq10}
\end{equation}
where $I_m^i$ represents the random mask resized to the shape of $AM_i$.

\begin{figure}[!t]
    \centering
    \includegraphics[scale=0.95]{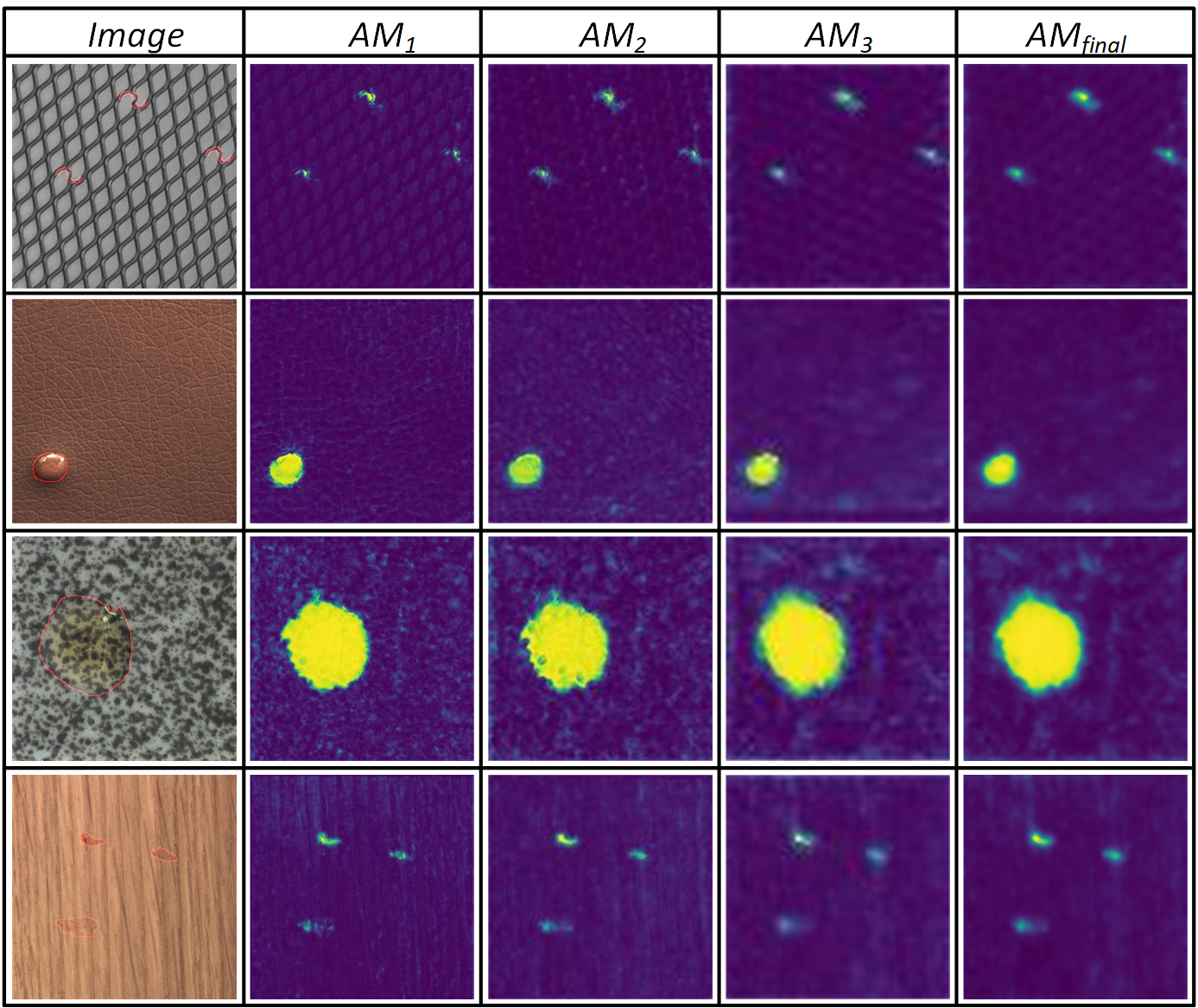}
    \caption{Visualization results of the Multi-Scale Defect Segmentation Module (MSDSM). All the above defective images are from the MVTec AD dataset \cite{MVTEC}.}
    \label{fig_7}
\end{figure}

Finally, some examples of the multiscale anomaly maps handled by the MSDSM module are shown in Fig. \ref{fig_7}. The anomaly maps at different scales have different discriminability for the defects. The $AM_{1}$ can capture more details of defects, but at the same time, it has a lot of noise. The $AM_{3}$ has a stronger discriminability for the defects; thus, it has less noise but lacks detailed information on the defects such as edges. Therefore, we apply the weighted sum of the anomaly maps at different scales to obtain the final anomaly map $AM_{final}$:
\begin{equation}
A{M_{final}} = {\lambda _1}A{M_1} + {\lambda _2}g(A{M_2}) + {\lambda _3}g(A{M_3})
\label{eq11}
\end{equation}
where $g(\cdot)$ denotes a bilinear up-sampling function that resizes the anomaly map to the size of the input image; and $\lambda_1$, $\lambda_2$ and $\lambda_3$ are the corresponding weights. In this paper, they are set as $\lambda_1$=0.2, $\lambda_2$=0.2, and $\lambda_3$=0.6. A 5
$\times$5 mean and median filters are applied to smooth the final anomaly map. The maximum value of $A{M_{final}}$ is utilized as the image-based anomaly score. 

\begin{figure}[!t]
    \centering
    \includegraphics{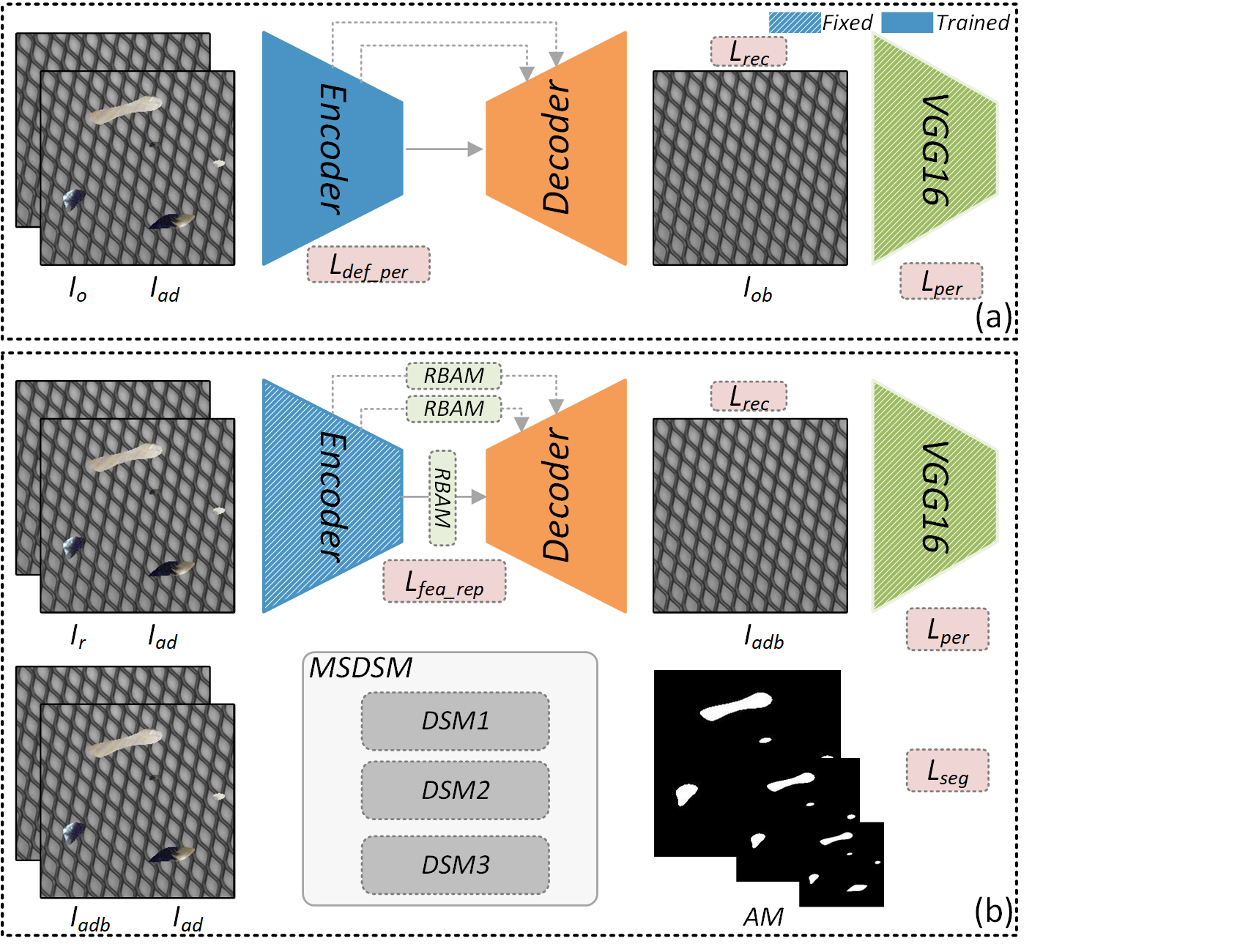}
    \caption{Illustration of the training and testing procedures of NDP-Net. (a) Training Phase 1. (b) Training Phase 2. The testing procedure of the NDP-Net is the same as in Training Phase 2. }
    \label{fig_8}
\end{figure}

\subsection{Training and Testing Procedures}
To make the NDP-Net have both a normal/abnormal discriminability and defect repair capability, a two-phase training strategy is applied to optimize the entire model.

As shown in Fig. \ref{fig_8} (a), the learning targets for Training Phase 1 are background reconstruction and pixel-level defect perception in the encoder. During Training Phase 1, only the encoder and decoder are optimized. Regarding
texture background reconstruction, the purpose of the loss function is to minimize the distance between the input image and the corresponding reconstructed image. Therefore, we simply use the mean square errors as the reconstruction loss:
\begin{equation}
  L_{rec}^1 = \underset{I_o \sim P_{I_{o}}}{\mathbb{E} }\left [ {||I_{o}-I_{ob}||}^2 \right ] 
  \label{eq12}
\end{equation}
where ${||\cdot||}^2$ denotes the $L_{2}$ norm.
To capture the high-level semantic information, the perceptual loss $L_{per}^1$ is leveraged by calculating the $L_{1}$ distance between the feature maps of $I_{o}$ and $I_{ob}$:
\begin{equation}
       L_{per}^1 = \underset{I_o \sim P_{I_{o}}}{\mathbb{E}} \left[\frac{1}{L}  \sum_{l=1}^{L} {||D_l(I_o)-D_l(I_{ob})||}_1 \right]
       \label{eq13}
\end{equation}
where $D_l$ denotes the $l$th feature map from the ImageNet-pretrained VGG-16 backbone. We utilize the first five feature maps from the ReLU activation layers; thus, $L$ is set to 5.

As mentioned in Section \MakeUppercase{\romannumeral3}-B, the encoder is trained with a pixel-level defect perception loss $L_{def\_per}$ to enable the model to be equipped with a pixel-level defect perception ability.

We can obtain the weighted joint loss for Training Phase 1 by combining Eqs. (\ref{eq2}), (\ref{eq12}) and (\ref{eq13}):
\begin{equation}
    L_{phase1}=w_{rec}^1 L_{rec}^1+w_{per}^1 L_{per}^1 + w_{def\_per}^1 L_{def\_per}
    \label{eq14}
\end{equation}
where $w_{rec}^1$, $w_{per}^1$ and $w_{def\_per}^1$ are the weights of the three types of losses, and we empirically set $w_{rec}^1$=100, $w_{per}^1$=1 and $w_{def\_per}^1$=1. After the accomplishment of Training Phase 1, the model is equipped with a strong background reconstruction and pixel-level defect perception ability.

\begin{table*}[!t]\centering
\caption{Details of the texture surface dataset}
\label{table1}
\begin{tabular}{cccccccccccccccccc}
\Xhline{1.5pt}
\multicolumn{2}{c}{{\color[HTML]{333333} }}                                                   & \multicolumn{5}{c}{{\color[HTML]{333333} MVTec AD}}                                                                                                      & {\color[HTML]{333333} } & \multicolumn{10}{c}{{\color[HTML]{333333} DAGM}}                                                                                                                                                                                                                                                \\ \cline{3-7} \cline{9-18} 
\multicolumn{2}{c}{\multirow{-2}{*}{{\color[HTML]{333333} Texture Surface Dataset}}}          & {\color[HTML]{333333} Carpet} & {\color[HTML]{333333} Grid} & {\color[HTML]{333333} Leather} & {\color[HTML]{333333} Tile} & {\color[HTML]{333333} Wood} & {\color[HTML]{333333} } & {\color[HTML]{333333} C.1} & {\color[HTML]{333333} C.2} & {\color[HTML]{333333} C.3} & {\color[HTML]{333333} C.4} & C.5                        & C.6                        & C.7                        & C.8                        & C.9                        & C.10                       \\ \hline
{\color[HTML]{333333} }                           & {\color[HTML]{333333} W defects number}   & {\color[HTML]{333333} 0}      & {\color[HTML]{333333} 0}    & {\color[HTML]{333333} 0}       & {\color[HTML]{333333} 0}    & {\color[HTML]{333333} 0}    & {\color[HTML]{333333} } & {\color[HTML]{333333} 0}   & {\color[HTML]{333333} 0}   & {\color[HTML]{333333} 0}   & {\color[HTML]{333333} 0}   & 0                          & 0                          & 0                          & 0                          & 0                          & 0                          \\ \cline{2-18} 
\multirow{-2}{*}{{\color[HTML]{333333} Training}} & {\color[HTML]{333333} W/O defects number} & {\color[HTML]{333333} 280}    & {\color[HTML]{333333} 264}  & {\color[HTML]{333333} 245}     & {\color[HTML]{333333} 230}  & {\color[HTML]{333333} 247}  & {\color[HTML]{333333} } & {\color[HTML]{333333} 275} & {\color[HTML]{333333} 275} & {\color[HTML]{333333} 275} & {\color[HTML]{333333} 275} & {\color[HTML]{333333} 275} & {\color[HTML]{333333} 275} & {\color[HTML]{333333} 275} & {\color[HTML]{333333} 275} & {\color[HTML]{333333} 275} & {\color[HTML]{333333} 275} \\ \hline
{\color[HTML]{333333} }                           & {\color[HTML]{333333} W defects number}   & {\color[HTML]{333333} 88}     & {\color[HTML]{333333} 57}   & {\color[HTML]{333333} 72}      & {\color[HTML]{333333} 84}   & {\color[HTML]{333333} 60}   & {\color[HTML]{333333} } & {\color[HTML]{333333} 150} & {\color[HTML]{333333} 150} & {\color[HTML]{333333} 150} & {\color[HTML]{333333} 150} & 150                        & 150                        & 300                        & 300                        & 300                        & 300                        \\ \cline{2-18} 
\multirow{-2}{*}{{\color[HTML]{333333} Testing}}  & {\color[HTML]{333333} W/O defects number} & {\color[HTML]{333333} 21}     & {\color[HTML]{333333} 21}   & {\color[HTML]{333333} 32}      & {\color[HTML]{333333} 33}   & {\color[HTML]{333333} 19}   & {\color[HTML]{333333} } & {\color[HTML]{333333} 0}   & {\color[HTML]{333333} 0}   & {\color[HTML]{333333} 0}   & {\color[HTML]{333333} 0}   & 0                          & 0                          & 0                          & 0                          & 0                          & 0                          \\ \Xhline{1.5pt}
\end{tabular}
\end{table*}

\begin{table*}[]\centering
\caption{The image level ROCAUC results of different SOTA methods on five types of textures in MVTEC AD dataset}
\label{table2}
\begin{threeparttable}
\begin{tabular}{cccccccclcccc}
\Xhline{1.5pt}
                           & \multicolumn{7}{c}{Reconstruction-based methods}                                                                                                                                                    &  & \multicolumn{3}{c}{Embedding-based methods}                                     &                                        \\ \cline{2-8} \cline{10-12}
\multirow{-2}{*}{Category} & AE\_SSIM & AnoGAN & {\color[HTML]{333333} f-AnoGAN} & {\color[HTML]{333333} GANomaly} & {\color[HTML]{333333} VAE} & {\color[HTML]{333333} TrustMAE}       & {\color[HTML]{333333} RIAD}            &  & {\color[HTML]{333333} MKD}   & SPADE       & {\color[HTML]{333333} Pacth SVDD}  & \multirow{-2}{*}{\textbf{NDP-Net}}       \\ \hline
Carpet                     & 67.00    & 49.00  & {\color[HTML]{333333} 56.57}    & {\color[HTML]{333333} 84.20}    & 67.00                      & {\color[HTML]{333333} \textbf{97.43}} & {\color[HTML]{333333} 84.20}           &  & {\color[HTML]{333333} 79.30} & 92.80       & {\color[HTML]{333333} 92.90}       & {\color[HTML]{333333} {\ul 95.41}}     \\
Grid                       & 69.00    & 51.00  & {\color[HTML]{333333} 59.63}    & {\color[HTML]{333333} 74.30}    & 83.00                      & {\color[HTML]{333333} 99.08}          & {\color[HTML]{333333} \textbf{99.60}}  &  & {\color[HTML]{333333} 78.10} & 47.30       & {\color[HTML]{333333} 94.60}       & {\color[HTML]{333333} {\ul 99.37}}     \\
Leather                    & 46.00    & 52.00  & {\color[HTML]{333333} 62.50}    & {\color[HTML]{333333} 79.20}    & 71.00                      & {\color[HTML]{333333} 95.07}          & {\color[HTML]{333333} \textbf{100.00}} &  & {\color[HTML]{333333} 95.10} & {\ul 95.40} & {\color[HTML]{333333} 90.90}       & {\color[HTML]{333333} \textbf{100.00}} \\
Tile                       & 52.00    & 51.00  & {\color[HTML]{333333} 61.34}    & {\color[HTML]{333333} 78.50}    & 81.00                      & {\color[HTML]{333333} 97.29}          & {\color[HTML]{333333} {\ul 98.70}}     &  & {\color[HTML]{333333} 91.60} & 96.50       & {\color[HTML]{333333} 97.80}       & {\color[HTML]{333333} \textbf{99.44}}  \\
Wood                       & 83.00    & 68.00  & {\color[HTML]{333333} 75.00}    & {\color[HTML]{333333} 65.30}    & 89.00                      & {\color[HTML]{333333} \textbf{99.82}} & {\color[HTML]{333333} 93.00}           &  & {\color[HTML]{333333} 94.30} & 95.80       & {\color[HTML]{333333} {\ul 96.50}} & {\color[HTML]{333333} \textbf{99.82}}  \\ \hline
Average                    & 64.40    & 52.20  & {\color[HTML]{333333} 63.01}    & {\color[HTML]{333333} 76.30}    & 78.20                      & {\ul 97.74}                           & 95.10                                  &  & 87.68                        & 85.56       & 94.54                              & {\color[HTML]{333333} \textbf{98.81}}  \\ \Xhline{1.5pt}
\end{tabular}
\begin{tablenotes}
\footnotesize
\item[1] The best image level ROCAUC performance is indicated by bold font, while the second best is indicated by an underline.
\end{tablenotes}
\end{threeparttable}
\end{table*}

\begin{table*}[]\centering
\caption{The pixel level ROCAUC results of different SOTA methods on five types of textures in MVTEC AD dataset}
\label{table3}
\begin{threeparttable}
\begin{tabular}{cccccclcccccc}
\Xhline{1.5pt}
                           & \multicolumn{5}{c}{Reconstruction-based methods}                                                                            &  & \multicolumn{5}{c}{Embedding-based methods}                                                       &                                       \\ \cline{2-6} \cline{8-12}
\multirow{-2}{*}{Category} & AE-SSIM & AnoGAN & {\color[HTML]{333333} VAE} & {\color[HTML]{333333} TrustMAE}    & {\color[HTML]{333333} RIAD}           &  & {\color[HTML]{333333} MKD}   & SPADE & DFR   & GCPF           & {\color[HTML]{333333} Pacth SVDD} & \multirow{-2}{*}{\textbf{NDP-Net}}      \\ \hline
Carpet                     & 87.00   & 54.00   & 73.50                      & {\color[HTML]{333333} {\ul 98.53}} & {\color[HTML]{333333} 96.30}          &  & {\color[HTML]{333333} 95.60} & 97.50 & 97.00 & \textbf{98.90} & {\color[HTML]{333333} 92.60}      & {\color[HTML]{333333} 97.62}          \\
Grid                       & 94.00   & 58.00   & 96.10                      & {\color[HTML]{333333} 97.45}       & {\color[HTML]{333333} \textbf{98.80}} &  & {\color[HTML]{333333} 91.80} & 93.70 & 98.00 & 97.80          & {\color[HTML]{333333} 96.20}      & {\color[HTML]{333333} {\ul 98.50}}    \\
Leather                    & 78.00   & 64.00   & 92.50                      & {\color[HTML]{333333} 98.05}       & {\color[HTML]{333333} {\ul 99.40}}    &  & {\color[HTML]{333333} 98.10} & 97.60 & 98.00 & 99.30          & {\color[HTML]{333333} 97.40}      & {\color[HTML]{333333} \textbf{99.43}} \\
Tile                       & 59.00   & 50.00   & 65.40                      & {\color[HTML]{333333} 82.48}       & {\color[HTML]{333333} 89.10}          &  & {\color[HTML]{333333} 82.80} & 87.40 & 87.00 & {\ul 96.10}    & {\color[HTML]{333333} 91.40}      & {\color[HTML]{333333} \textbf{99.52}} \\
Wood                       & 73.00   & 62.00   & 83.80                      & {\color[HTML]{333333} 92.62}       & {\color[HTML]{333333} 85.80}          &  & {\color[HTML]{333333} 84.80} & 88.50 & 94.00 & {\ul 95.10}    & {\color[HTML]{333333} 90.80}      & {\color[HTML]{333333} \textbf{97.64}} \\ \hline
Average                    & 78.00   & 58.00   & 82.26                      & 93.83                              & 93.70                                 &  & 90.62                        & 92.94 & 94.80 & {\ul 97.44}    & 93.68                             & {\color[HTML]{333333} \textbf{98.54}} \\ \Xhline{1.5pt}
\end{tabular}
\begin{tablenotes}
\footnotesize  
\item[1] The best pixel level ROCAUC performance is indicated by bold font, while the second best is indicated by an underline.
\end{tablenotes}
\end{threeparttable}
\end{table*}

\begin{table*}[]\centering
\caption{The PROAUC results of different SOTA methods on five types of textures in MVTEC AD dataset}
\label{table4}
\begin{threeparttable}
\begin{tabular}{ccccccccccccc}
\Xhline{1.5pt}
                           & \multicolumn{4}{c}{Reconstruction-based methods}                &  & \multicolumn{6}{c}{Embedding-based methods}                                                      &                                       \\ \cline{2-5} \cline{7-12}
\multirow{-2}{*}{Category} & AE\_SSIM & AnoGAN & {\color[HTML]{333333} VAE} & EdgRec         &  & SPADE & {\color[HTML]{333333} CNN\_Dict} & ST    & MB-PFM         & PaDiM       & STPM           & \multirow{-2}{*}{\textbf{NDP-Net}}      \\ \hline
Carpet                     & 64.70    & 20.40  & 50.10                      & \textbf{96.90} &  & 94.70 & {\color[HTML]{333333} 46.90}     & 87.90 & \textbf{96.90} & {\ul 96.20} & 95.80          & {\color[HTML]{333333} 93.20}          \\
Grid                       & 84.90    & 22.60  & 22.40                      & \textbf{97.00} &  & 86.70 & {\color[HTML]{333333} 18.30}     & 95.20 & 96.00          & 94.60       & {\ul 96.60}    & {\color[HTML]{333333} 95.72}          \\
Leather                    & 56.10    & 37.80  & 63.50                      & \textbf{98.80} &  & 97.20 & {\color[HTML]{333333} 64.10}     & 94.50 & \textbf{98.80} & 97.80       & 98.00          & {\color[HTML]{333333} {\ul 98.79}}    \\
Tile                       & 17.50    & 17.70  & 87.00                      & {\ul 96.30}    &  & 75.90 & {\color[HTML]{333333} 79.70}     & 94.60 & 88.70          & 86.00       & 92.10          & {\color[HTML]{333333} \textbf{97.39}} \\
Wood                       & 60.50    & 38.60  & 62.80                      & 77.50          &  & 87.40 & {\color[HTML]{333333} 62.10}     & 91.10 & 92.60          & 91.10       & \textbf{93.60} & {\color[HTML]{333333} {\ul 93.55}}    \\ \hline
Average                    & 56.74    & 27.42  & 57.16                      & 93.30          &  & 88.38 & 54.22                            & 92.70 & 94.60          & 93.14       & {\ul 95.22}    & {\color[HTML]{333333} \textbf{95.73}} \\ \Xhline{1.5pt}
\end{tabular}
\begin{tablenotes}
\footnotesize
\item[1] The best PROAUC performance is indicated by bold font, while the second best is indicated by an underline.
\end{tablenotes}
\end{threeparttable}
\end{table*}

As shown in Fig. \ref{fig_8} (b), to avoid affecting the defect perception ability, the weights of the encoder are no longer updated. The learning targets for Training Phase 2 are defect restoration and accurate defect segmentation. During Training Phase 2, the RBAM, the decoder, and the MSDSM are optimized together. For defect restoration, the purpose of the loss function is to minimize the distance between the reconstructed image ($I_{adb}$) of the artificial defect and the source defect-free image ($I_{o}$). In addition, perceptual loss is also applied to generate reconstructed images that are more in line with human perception.
Thus, $L_{rec}^2$ and $L_{per}^2$ are defined as:
\begin{equation}
      L_{rec}^2 = \underset{I_o \sim P_{I_{o}}}{\mathbb{E} }\left [ {||I_{o}-I_{adb}||}^2 \right ] 
      \label{eq15}
\end{equation}
\begin{equation}
    L_{per}^2 = \underset{I_o \sim P_{I_{o}}}{\mathbb{E}} \left[\frac{1}{L}  \sum_{l=1}^{L} {||D_l(I_o)-D_l(I_{adb})||}_1 \right]
    \label{eq16}
\end{equation}

As mentioned in Section \MakeUppercase{\romannumeral3}-C, the feature repair loss ($L_{fea\_rep}$) is utilized to guide the RBAM to learn how to convert the defective features to normal features.

For accurate defect segmentation, as mentioned in Section \MakeUppercase{\romannumeral3}-D, the segmentation loss ($L_{seg}$) is leveraged to improve the robustness of the MSDSM.

With the combination of Eqs. (\ref{eq9}), (\ref{eq10}), (\ref{eq15}) and (\ref{eq16}), the weighted joint loss function for Training Phase 2 can be defined as:
\begin{equation}
\begin{split}
    L_{phase2}=w_{rec}^2 L_{rec}^2+w_{per}^2 L_{per}^2 +w_{fea\_rep}^2 L_{fea\_rep}\\ 
    +w_{seg}^2 L_{seg}
\end{split}
\label{eq17}
\end{equation}
where $w_{rec}^2$, $w_{per}^2$, $w_{fea\_rep}^2$ and $w_{seg}^2$ are the weights of the four types of losses, and they are set as $w_{rec}^2$=100, $w_{per}^2$=1, $w_{fea\_rep}^2$=1 and $w_{seg}^2$=1. After the accomplishment of Training Phase 2, the NDP-Net model can achieve accurate surface defect detection.

During the testing procedure, similar to Training Phase 2, when the defective image is inputted, the NDP-Net model can address the defective features to suppress the reconstruction of defects. Then, the input defective image and the defect-free reconstructed image are reinput to the encoder to obtain multi-scale concatenated features. Finally, MSDSM utilizes the multi-scale concatenated features to obtain the final anomaly map.

\section{Experimental Results}
To evaluate the performance of the proposed NDP-Net method, a series of experiments are conducted in this section. First, the datasets and implementation details are introduced. Second, the overall detection performance of NDP-Net on MVTec AD \cite{MVTEC} is compared with that of several state-of-the-art methods. Third, a comparative experiment on the DAGM \cite{DAGM} dataset is conducted to further demonstrate the generalization of the NDP-Net. Fourth, an ablation study is conducted to illustrate the influence of each module in the NDP-Net, including $L_{def\_per}$, RBAM, and MSDSM. Finally, the industrial application of NDP-Net is introduced.

\subsection{Datasets and Implementation Details}
\subsubsection{Datasets}
The MVTec AD dataset \cite{MVTEC} has been widely used for unsupervised defect detection and segmentation and consists of 5 types of textures: carpet, grid, leather, tile, and wood. A total of 1266 defect-free texture images and no defective images are given for training. There are 126 defect-free images and 361 defective images for testing. The texture defects in the testing images are variant in the color, scale, shape, etc., which makes MVTec AD very challenging. Furthermore, to demonstrate the good generalization of the model, we also verify the performance of the NDP-Net on the DAGM \cite{DAGM} dataset, which consists of 10 types of irregular textures. The defects vary in shape and size, which makes it difficult to locate the defective regions on complex irregular texture surfaces. In our experiments, each image is resized to a resolution of 256$\times$256. More details of the datasets are summarized in TABLE \ref{table1}.

\subsubsection{Implementation Details and Evaluation Criterion}
All experiments are implemented on a computer with Xeon(R) Silver 4116 CPU@2.10 GHz and NVIDIA V100 GPU with a 32-GB memory size. The entire NDP-Net model is optimized using the Adam optimizer with a learning rate of 0.0001 and a weight decay of 0.00001. The training epoch sizes for the two-phase training procedure are set to 400 and 200.

The key to fairly comparing the performance of each method is the evaluation criterion. In this paper, we follow the area under the receiver operating characteristic curve (ROCAUC) for detection and segmentation, which is commonly used as the evaluation metric in unsupervised anomaly detection. However, the ROCAUC for segmentation favors large defective areas, which leads to an inaccurate evaluation. Thus, we also adopt the normalized area under the per-region overlap curve (PROAUC) as the defect segmentation evaluation criterion.
\begin{figure}[!t]
    \centering
    \includegraphics[scale=0.98]{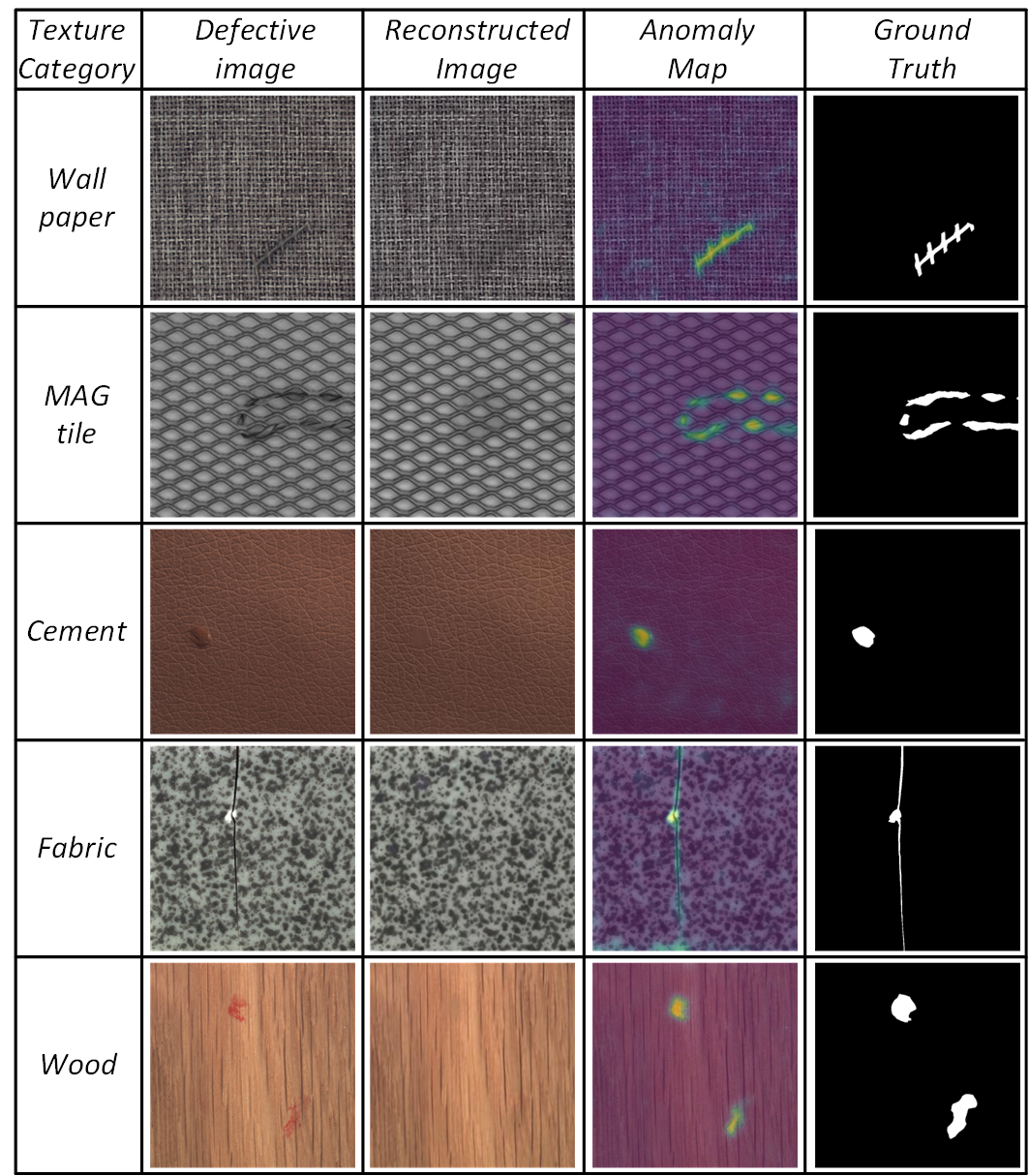}
    \caption{Reconstruction and segmentation results on the various texture images. All the above defective images are from the MVTec AD dataset \cite{MVTEC}.}
    \label{fig_9}
\end{figure}
\begin{figure}[!t]
    \centering
    \includegraphics[scale=0.98]{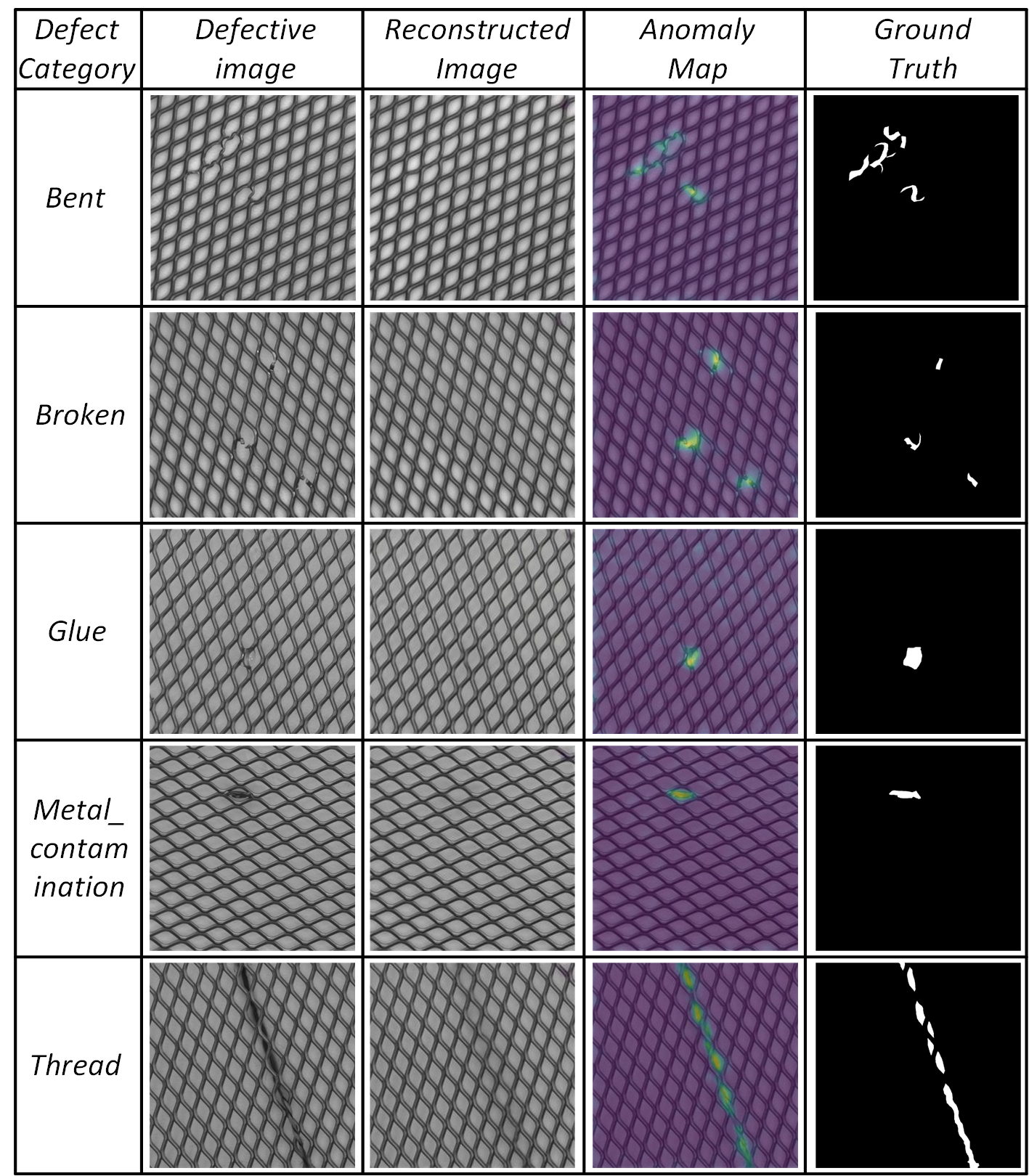}
    \caption{Reconstruction and segmentation results on the various types of defects. All the above defective images are from the MVTec AD dataset \cite{MVTEC}.}
    \label{fig_10}
\end{figure}
\begin{figure}[!t]
    \centering
    \includegraphics[scale=1.0]{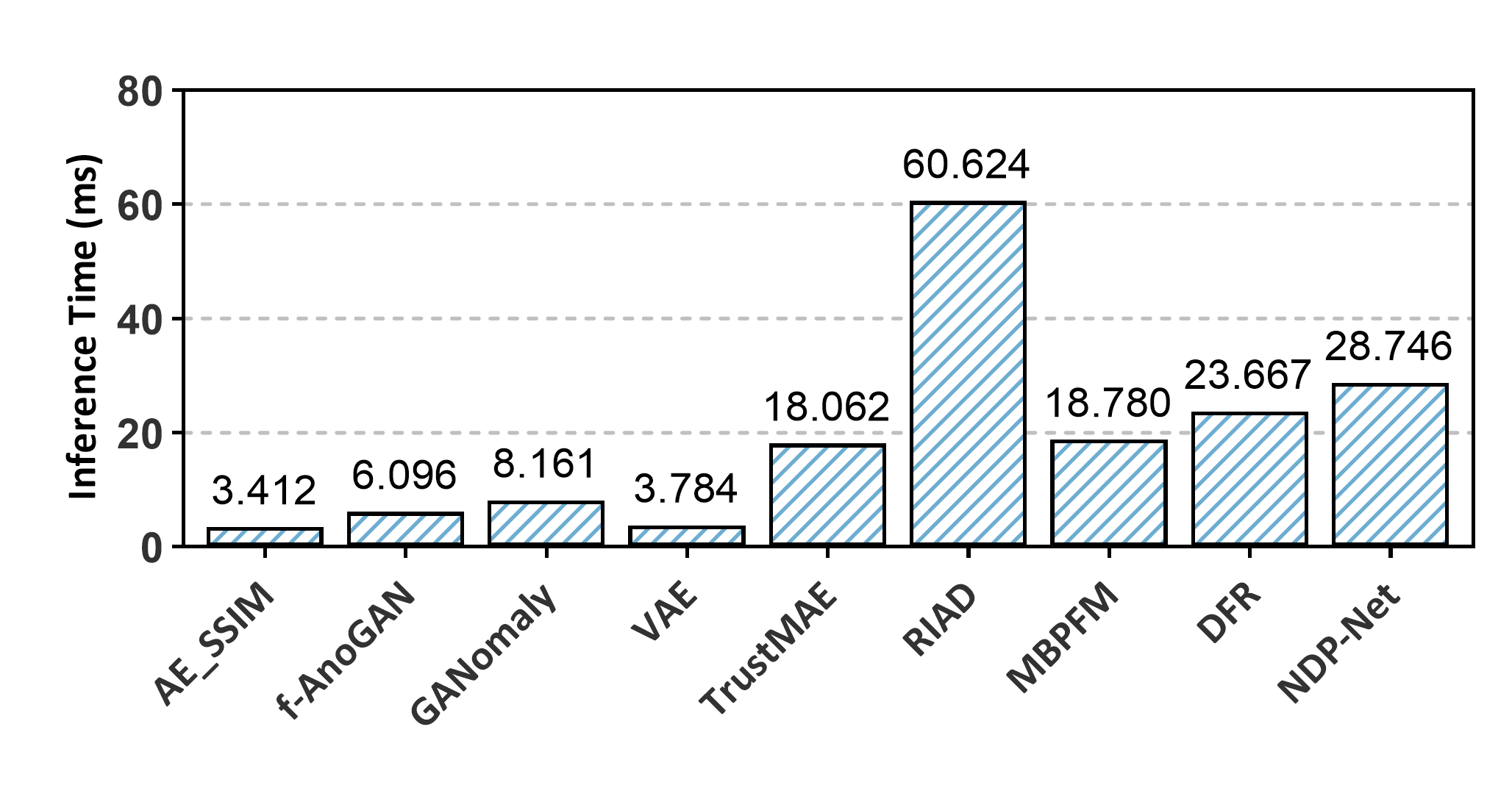}
    \caption{Comparison of inference time.}
    \label{fig_11}
\end{figure}
\subsection{Overall Comparative Experiments on MVTec AD}
To verify the performance of the proposed NDP-Net method, the inspection performance of NDP-Net is compared with the long-standing reconstruction-based methods AE\_SSIM \cite{AE-SSIM}, f-AnoGAN \cite{f-anogan}, AnoGAN \cite{AnoGan}, GANomaly \cite{Ganomaly}, VAE \cite{VAE}, TrustMAE \cite{TrustMAE}, RIAD \cite{RIAD}, EdgRec
\cite{edgrec}, and the embedding-based methods MKD \cite{MKD}, CNN\_Dict \cite{CNN_Dict}, SPADE \cite{SPADE}, DFR \cite{DFR}, GCPF \cite{GCPF}, Patch SVDD \cite{PatchSVDD}, ST \cite{ST}, MB-PFM \cite{MBPFM}, PaDiM \cite{PaDiM}, and STPM \cite{STPM}.
\subsubsection{Defect Detection}
As shown in TABLE \ref{table2}, the proposed NDP-Net achieves a better defect detection performance than that of other state-of-the-art methods. AE\_SSIM, AnoGAN, and f-AnoGAN do not perform well. Although RIAD, TrustMAE, and Patch SVDD perform better, they are unable to maintain good inspection performance for all texture categories. For example, RIAD achieves outstanding performance on the leather dataset with an image level ROCAUC of 100.00\% but poor performance on the carpet dataset with an image level ROCAUC of 84.20\%. In contrast, our proposed NDP-Net can perform well for all categories. This is because $L_{def\_per}$ makes the extracted features more discriminative, and the MSDSM utilizes these more discriminative features to easily detect defects.

\begin{table*}[!t]\centering
\caption{Pixel-level ROCAUC of different methods on ten types of texture in the DAGM dataset}
\label{table5}
\begin{threeparttable}
\begin{tabular}{ccccccccccc|c|c}
\Xhline{1.5pt}
{\color[HTML]{333333} Category}       & {\color[HTML]{333333} Class1}         & {\color[HTML]{333333} Class2}         & {\color[HTML]{333333} Class3}         & {\color[HTML]{333333} Class4}         & {\color[HTML]{333333} Class5}         & {\color[HTML]{333333} Class6}         & {\color[HTML]{333333} Class7}         & {\color[HTML]{333333} Class8}         & {\color[HTML]{333333} Class9}         & {\color[HTML]{333333} Class10}        & {\color[HTML]{333333} Average}        & {\color[HTML]{333333} Standard Deviation} \\ \hline
{\color[HTML]{333333} AE\_SSIM}       & {\color[HTML]{333333} 68.50}          & {\color[HTML]{333333} 73.60}          & {\color[HTML]{333333} 74.30}          & {\color[HTML]{333333} 87.80}          & {\color[HTML]{333333} 81.30}          & {\color[HTML]{333333} 79.40}          & {\color[HTML]{333333} 95.60}          & {\color[HTML]{333333} 68.10}          & {\color[HTML]{333333} 97.40}          & {\color[HTML]{333333} 81.70}          & {\color[HTML]{333333} 80.77}          & 9.765                                     \\ \hline
{\color[HTML]{333333} CNN\_Dict}      & {\color[HTML]{333333} 88.70}          & {\color[HTML]{333333} 58.10}          & {\color[HTML]{333333} 69.20}          & {\color[HTML]{333333} 68.20}          & {\color[HTML]{333333} 57.30}          & {\color[HTML]{333333} 75.80}          & {\color[HTML]{333333} 66.50}          & {\color[HTML]{333333} 59.60}          & {\color[HTML]{333333} 65.10}          & {\color[HTML]{333333} 67.90}          & {\color[HTML]{333333} 67.64}          & 8.850                                     \\ \hline
{\color[HTML]{333333} AnoGAN}         & {\color[HTML]{333333} 21.80}          & {\color[HTML]{333333} 66.60}          & {\color[HTML]{333333} 78.10}          & {\color[HTML]{333333} 86.30}          & {\color[HTML]{333333} 70.10}          & {\color[HTML]{333333} 45.50}          & {\color[HTML]{333333} 72.20}          & {\color[HTML]{333333} 60.90}          & {\color[HTML]{333333} 94.60}          & {\color[HTML]{333333} 72.70}          & {\color[HTML]{333333} 66.88}          & 19.670                                    \\ \hline
{\color[HTML]{333333} OCGAN}          & {\color[HTML]{333333} 18.70}          & {\color[HTML]{333333} 86.60}          & {\color[HTML]{333333} 89.10}          & {\color[HTML]{333333} 62.30}          & {\color[HTML]{333333} 91.80}          & {\color[HTML]{333333} 74.50}          & {\color[HTML]{333333} 97.10}          & {\color[HTML]{333333} 78.10}          & {\color[HTML]{333333} 90.30}          & {\color[HTML]{333333} 95.60}          & {\color[HTML]{333333} 78.41}          & 22.332                                    \\ \hline
{\color[HTML]{333333} MS-FCAE}        & {\color[HTML]{333333} 96.20}          & {\color[HTML]{333333} 80.70}          & {\color[HTML]{333333} 95.90}          & {\color[HTML]{333333} 87.40}          & {\color[HTML]{333333} 94.00}          & {\color[HTML]{333333} 58.80}          & {\color[HTML]{333333} 90.50}          & {\color[HTML]{333333} 73.90}          & {\color[HTML]{333333} 96.80}          & {\color[HTML]{333333} 96.40}          & {\color[HTML]{333333} 87.06}          & 11.900                                    \\ \hline
{\color[HTML]{333333} AFEAN}          & {\color[HTML]{333333} 97.10}          & {\color[HTML]{333333} 94.90}          & {\color[HTML]{333333} 97.40}          & {\color[HTML]{333333} 89.50}          & {\color[HTML]{333333} 96.70}          & {\color[HTML]{333333} 94.80}          & {\color[HTML]{333333} 98.30}          & {\color[HTML]{333333} 81.30}          & {\color[HTML]{333333} 98.80}          & {\color[HTML]{333333} 96.10}          & {\color[HTML]{333333} 94.49}          & 5.050                                     \\ \hline
{\color[HTML]{333333} ACDN}           & {\color[HTML]{333333} {\ul 99.20}}    & {\color[HTML]{333333} {\ul 96.80}}    & {\color[HTML]{333333} {\ul 99.30}}    & {\color[HTML]{333333} {\ul 92.80}}    & {\color[HTML]{333333} {\ul 97.60}}    & {\color[HTML]{333333} {\ul 97.90}}    & {\color[HTML]{333333} {\ul 99.80}}    & {\color[HTML]{333333} {\ul 86.50}}    & {\color[HTML]{333333} {\ul 99.60}}    & {\color[HTML]{333333} {\ul 96.70}}    & {\color[HTML]{333333} {\ul 96.62}}    & {\ul 3.905}                               \\ \hline
{\color[HTML]{333333} \textbf{NDP-Net}} & {\color[HTML]{333333} \textbf{99.80}} & {\color[HTML]{333333} \textbf{99.22}} & {\color[HTML]{333333} \textbf{99.94}} & {\color[HTML]{333333} \textbf{98.74}} & {\color[HTML]{333333} \textbf{99.76}} & {\color[HTML]{333333} \textbf{99.42}} & {\color[HTML]{333333} \textbf{99.94}} & {\color[HTML]{333333} \textbf{99.89}} & {\color[HTML]{333333} \textbf{99.64}} & {\color[HTML]{333333} \textbf{99.50}} & {\color[HTML]{333333} \textbf{99.59}} & \textbf{0.361}                            \\ \Xhline{1.5pt}
\end{tabular}
\begin{tablenotes}
\footnotesize  
\item[1] The best pixel level ROCAUC performance/Standard Deviation is indicated by bold font, while the second best is indicated by an underline.
\end{tablenotes}
\end{threeparttable}
\end{table*}

\begin{figure*}[!t]
    \centering
    \includegraphics[scale=0.95]{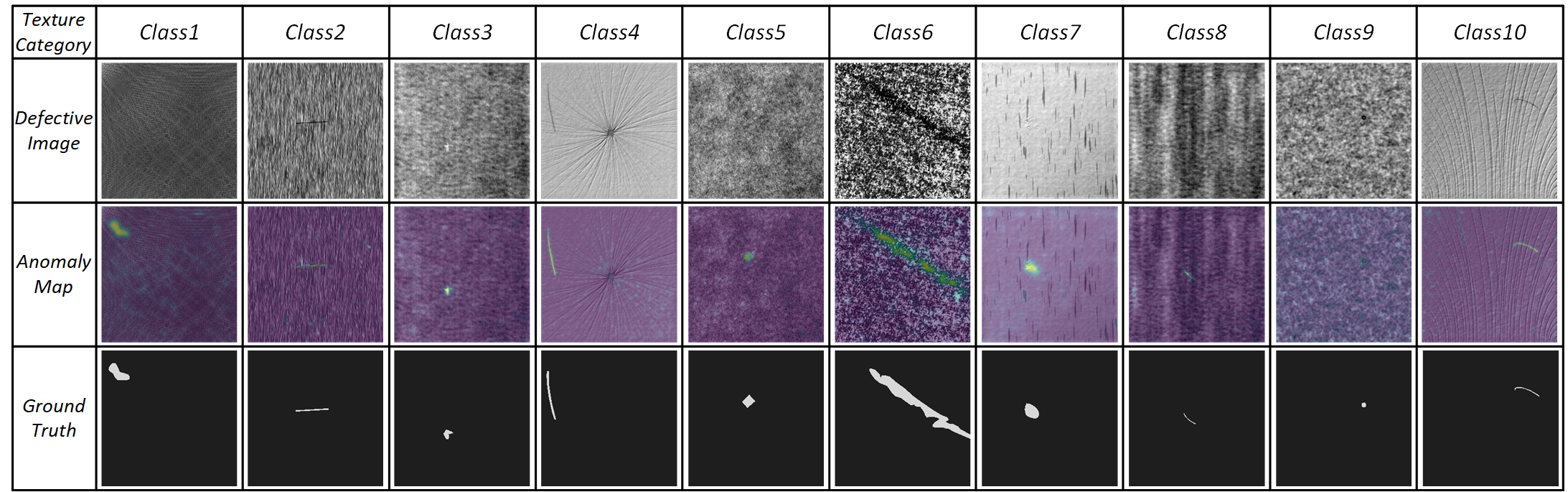}
    \caption{Segmentation results on the various texture images of DAGM \cite{DAGM}.}
    \label{fig_12}
\end{figure*}
\subsubsection{Defect Segmentation}
As shown in TABLE \ref{table3}, the NDP-Net also achieves better defect segmentation performance compared with other outstanding methods. Especially for Tile and Wood, the NDP-Net improves the pixel-level ROCAUC by margins of 3.42\% and 2.54\% compared to the second-best results. This demonstrates that the NDP-Net can detect defects well at the pixel level. However, because the pixel ROCAUC favors the large defective areas, which may lead to the inaccurate evaluation for defect segmentation, the PROAUC criterion is also utilized to evaluate the performance of defect segmentation. As shown in TABLE \ref{table4}, the NDP-Net achieves a better average result of overall texture categories on the defect segmentation. Compared to the second-best reconstruction-based method, the NDP-Net improves the PROAUC by margins of 2.43\%. This is because RBAM can suppress the reconstruction of defects well, and thus, we can obtain defect-free reconstruction images, which contributes to a more accurate defect localization.

\subsubsection{Inference Time}
We assume that an industry-compliant method needs to achieve a good balance between the inference time and overall inspection performance. To verify that our method can meet industrial inspection demands, a comparative experiment on the inference time is conducted. The inference time of NDP-Net compared with other long-standing methods is shown in Fig. \ref{fig_11}. The inference time of NDP-Net is 28.746. Although the inference time of NDP-Net lags significantly behind AE\_SSIM, f-AnoGAN, GANomaly, and VAE, the NDP-Net performs much better than the above methods in terms of the inspection performance. The inference time of RIAD is too long, which limits its industrial application. TrustMAE, MBPFM, and DFR achieve a decent detection speed and have a seemingly good inspection performance in terms of average results. However, the robustness of these methods is poor. For example, as shown in TABLE \ref{table3}, TrustMAE achieves outstanding performance with a pixel-level ROCAUC of 98.53\% on the carpet dataset but poor performance with a pixel-level ROCAUC of 82.48\% on the tile dataset, which is a serious problem in real industrial scenarios. In contrast, our proposed NDP-Net performs well on all texture datasets, which reveals that NDP-Net has good robustness. Overall, the NDP-Net can achieve a good balance between the inference time and overall inspection performance, which is in line with industrial demands.
\subsubsection{Visualization Results}Some detection results of NDP-Net on the five types of texture images are shown in Fig. \ref{fig_9}. The NDP-Net can not only reconstruct the normal texture background and repair the defects but also achieve accurate defect segmentation on all types of texture surfaces. In addition, as shown in Fig. \ref{fig_10}, the NDP-Net is also capable of accurately inspecting the various types of defects, such as bent, broken, glue, metal\_contamination, and thread defects.

\subsection{Inspection Generalization Experiments on DAGM}
To demonstrate the generalization of NDP-Net, the inspection performance of NDP-Net on DAGM \cite{DAGM} is compared with the outstanding methods AE\_SSIM \cite{AE-SSIM}, CNN\_Dict \cite{CNN_Dict}, AnoGAN \cite{AnoGan}, OCGAN \cite{OCGAN}, MS-FCAE \cite{MSFCAE}, AFEAN \cite{AFEAN}, and ACDN \cite{ACDN}.

The quantitative experimental results are shown in TABLE \ref{table5}. Compared to other outstanding methods, the proposed NDP-Net achieves the best inspection performance
on all categories of DAGM \cite{DAGM}. The NDP-Net improves the average pixel level ROCAUC on the ten categories by a margin of 2.97\% compared to the second-best result. In addition, the standard deviation of NDP-Net over 10 texture categories is 0.361, which is much smaller than the other methods, revealing its excellent robustness.

Some detection results of NDP-Net on ten types of texture images are shown in Fig. \ref{fig_12}. The NDP-Net can achieve accurate defect segmentation on all types of texture images.

\begin{table}
\caption{NDP-Net ablation study on the wood dataset}
\label{table6}
\setlength{\tabcolsep}{3pt}
\begin{tabular}{p{\columnwidth}}
\centering
\includegraphics[width=0.9\columnwidth]{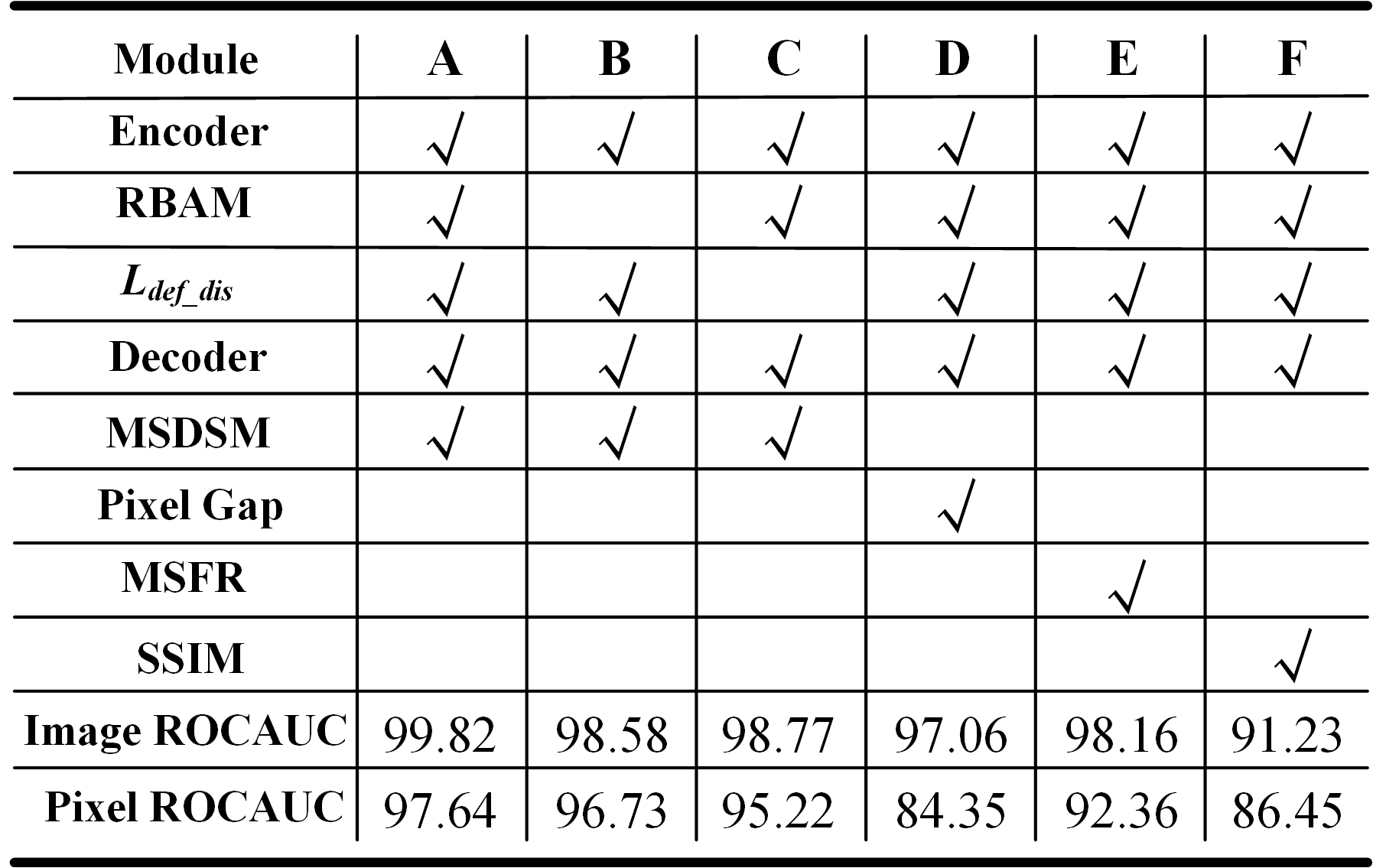}
\end{tabular}
\end{table}
\begin{figure}[!t]
    \centering
    \includegraphics[scale=0.95]{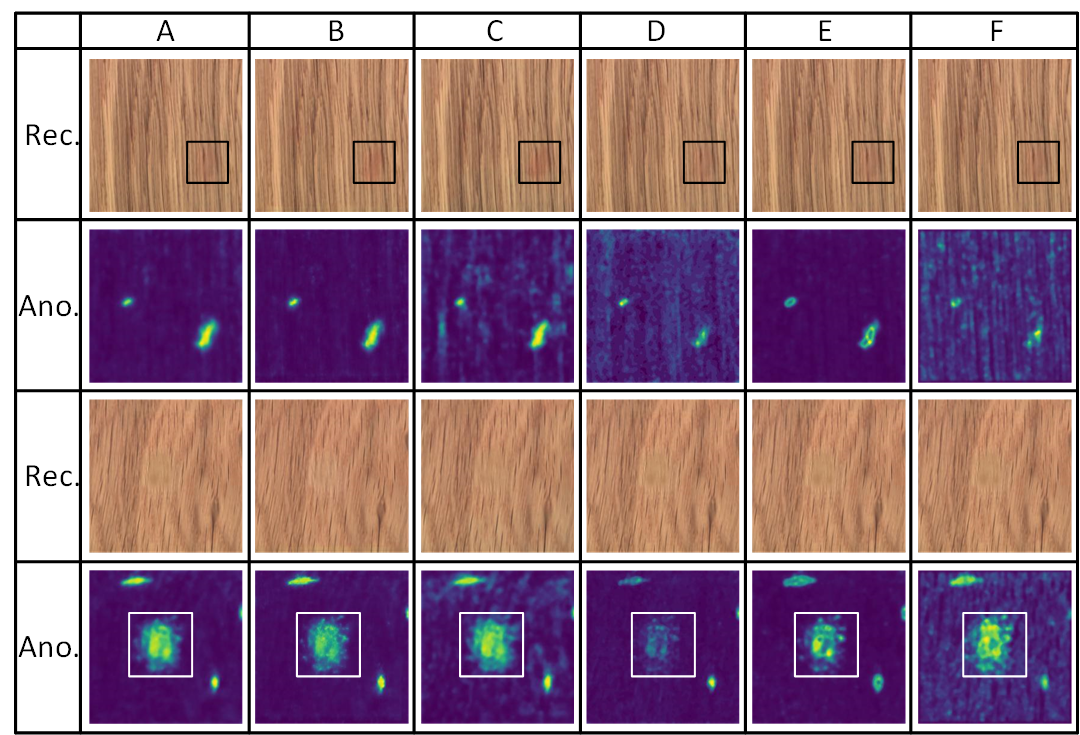}
    \caption{Examples of images from tests in the ablation study. Rec. denotes the reconstructed image and Ano. denotes the anomaly map.}
    \label{fig_13}
\end{figure}
\subsection{Ablation Study of the NDP-Net}
In this subsection, to analyze the effectiveness of each component of the NDP-Net, a series of ablation experiments are conducted on the wood dataset. In particular, we compare the MSDSM with other segmentation methods, such as the pixel gap between the input images and corresponding reconstructed images, multi-scale feature residual (MSFR) \cite{CMA-AE} and structure similarity index measure (SSIM) \cite{SSIM}. To guarantee a fair comparison, the parameter settings of all the variants of NDP-Net are maintained the same. The qualitative and quantitative experimental results are shown in Fig. \ref{fig_13} and TABLE \ref{table6}, respectively.
\subsubsection{Influence of $L_{def\_per}$}
$L_{def\_per}$ is a novel loss function, making it possible for the model to perceive normal/abnormal at the pixel level. The more discriminative features contribute to a more accurate defect detection and segmentation. Thus, the influence of $L_{def\_per}$ is analyzed in detail.

Column C of Fig. \ref{fig_13} shows examples of the influence of $L_{def\_per}$. The anomaly maps obtained by the model without $L_{def\_per}$ have more noise. This is because the model without $L_{def\_per}$ cannot perceive the defects at the pixel level. The quantitative results are described in Column C of TABLE \ref{table6}. When the NDP-Net is trained without $L_{def\_per}$, the image/pixel ROCAUC values decrease by 1.05/2.42\% compared to the entire NDP-Net (Column A).

These experimental results demonstrate that $L_{def\_per}$ can improve the overall inspection performance. Leveraging the $L_{def\_per}$ for training equips the model with the ability to distinguish normal/abnormal at the pixel level, which benefits defect detection and segmentation.

\begin{figure}[!t]
    \centering
    \includegraphics[scale=1]{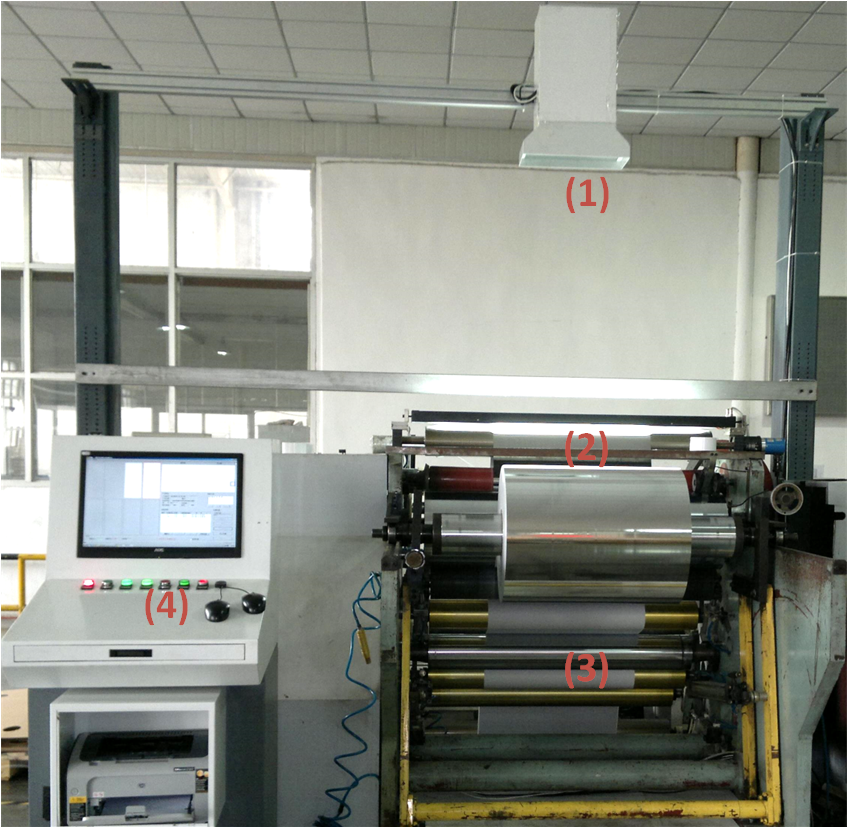}
    \caption{Automated optical inspection (AOI) equipment for package paper texture surface (PPTS) defect detection. (1) Camera. (2) Illumination system. (3) Package paper material. (4) software operation interface.}
    \label{fig_14}
\end{figure}
\subsubsection{Influence of RBAM}
The RBAM aims to leverage the reference normal features to repair the defective features, resulting in defect-free reconstructed images, which is the basis for subsequent defect segmentation. Therefore, the influence of RBAM is studied in detail.

Examples of the influence of RBAM are shown in Column B of Fig. \ref{fig_13}. Compared with the entire NDP-Net, the NDP-Net without RBAM cannot address the defective features well, which leads to mis-inspection. The quantitative results of NDP-Net without RBAM are shown in Column B of TABLE \ref{table6}. When the RBAM is removed, the image/pixel ROCAUC values decrease by 1.24/0.91\% compared to the entire NDP-Net (Column A).

These experimental results verify that the RBAM can improve the ability to address defects. Using the reference features to repair the defective features, the model can obtain clear reconstructed images, which is helpful to achieve more accurate defect segmentation.

\subsubsection{Influence of MSDSM}
The MSDSM is used to segment the defects through the input images and corresponding reconstructed images, which is the key to achieving accurate defect segmentation. Thus, the MSDSM is compared with the pixel gap, MSFR, and SSIM segmentation methods.

Columns D, E, and F of Fig. \ref{fig_13} show examples of the influence of the MSDSM. The MSDSM can obtain clearer anomaly detection than the pixel gap (Column D) and SSIM (Column F).
Compared with the MSFR (Column E), the MSDSM can achieve a more accurate defect localization. The quantitative results are shown in Columns D, E, and F of TABLE \ref{table6}. When the MSDSM is replaced with a pixel gap, MSFR, and SSIM, the image/pixel ROCAUC values decrease by 2.76/13.29\%, 1.66/5.28\%, and 8.59/11.19\%, respectively, compared to the entire NDP-Net (Column A).

These experimental results demonstrate that the MSDSM is superior to other segmentation methods. MSDSM leverages the multi-scale concatenated discriminative features to segment defects, which further improves the overall performance.

\begin{figure}[!t]
    \centering
    \includegraphics[scale=0.98]{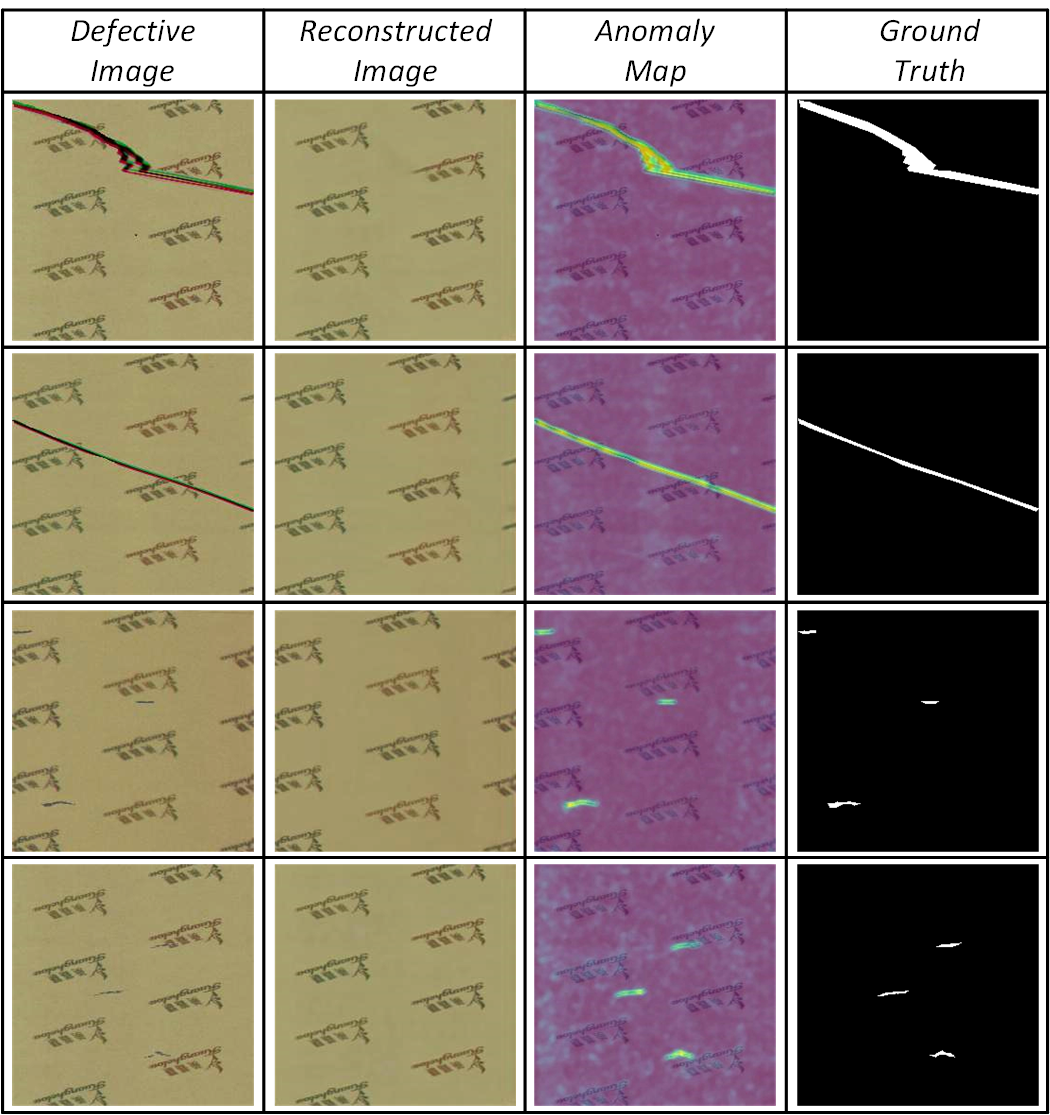}
    \caption{Some examples of the package paper texture surface (PPTS) defect inspection results.}
    \label{fig_15}
\end{figure}

\subsection{Industrial Application}
As shown in Fig. \ref{fig_14}, in addition to the open-source MVTec AD \cite{MVTEC} and DAGM \cite{DAGM} datasets, the proposed NDP-Net is implemented in our automated optical inspection equipment to online detect defects on the package paper texture surface (PPTS) dataset in real industrial scenarios. During the production process, a variety of defects occur on the textured surface. Therefore, it is impractical to collect all types of defects, which limits the application of supervised methods. In contrast, the unsupervised method NDP-Net proposed in this paper can accomplish accurate defect inspection in such industrial scenarios. These raw PPTS samples have a high resolution of 1913x2624 pixels. We extract small patches of 256x256 pixels from the raw PPTS samples to train the entire NDP-Net.

Some examples of PPTS defect inspection results are shown in Fig. \ref{fig_15}. The defective images, corresponding reconstructed images, anomaly maps, and ground truths are shown in the first, second, third, and fourth columns, respectively. As we can see, our proposed NDP-Net can not only reconstruct the normal texture backgrounds but also repair the defects, thus achieving accurate defect segmentation.

\section{Conclusion}
In this article, an unsupervised NDP-Net method is proposed for the accurate defect detection and segmentation. This method is only trained on defect-free images and corresponding artificial defective images. First, the pixel-level defect perception loss is leveraged to cause the feature extracted by the encoder to be more discriminative at the pixel level. In addition, a novel RBAM is proposed to suppress the reconstruction of defects, which leverages the normal feature of the fixed reference defect-free image to repair the defective feature. Then, the decoder utilizes the repaired features to reconstruct the texture background. Next, the reconstructed images and the original input images are reinputted into the encoder to obtain multi-scale concatenated features. Finally, the MSDSM utilizes the multi scale concatenated features to detect and segment the defects accurately. Extensive comparative experimental results on MVTec AD and DAGM demonstrate that the proposed NDP-Net achieves a state-of-the-art inspection and segmentation performance and has good generalization. In follow-up research, we will focus on how to apply this method to object surfaces.

\bibliographystyle{IEEEtran}
\bibliography{reference}

\begin{IEEEbiography}[{\includegraphics[width=1in,height=1.25in,clip,keepaspectratio]{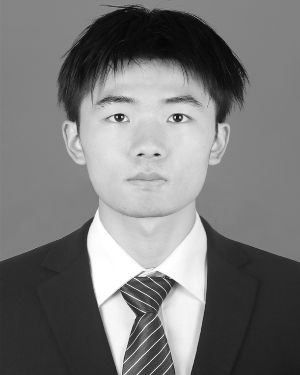}}]{Wei Luo}
will receive a B.S. degree from the School of Mechanical Science and Engineering, Huazhong University of Science and Technology, Wuhan, China, in 2023. He is gong to pursue a Ph.D. degree with the Department of Precision Instrument, Tsinghua University.

His research interests include deep learning, anomaly detection and machine vision.
\end{IEEEbiography}
\vspace{1\baselineskip}

\begin{IEEEbiography}[{\includegraphics[width=1.25in,height=1.25in,clip,keepaspectratio]{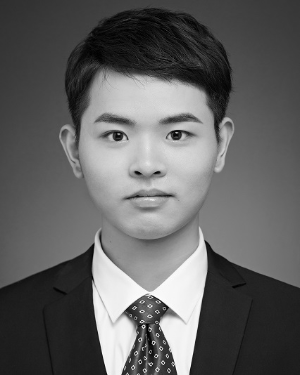}}]{Haiming Yao}
received a B.S. degree from the School of Mechanical Science and Engineering, Huazhong University of Science and Technology, Wuhan, China, in 2022. He is pursuing a Ph.D. degree with the Department of Precision Instrument, Tsinghua University.

His research interests include deep learning, edge intelligence and machine vision.
\end{IEEEbiography}
\vspace{1\baselineskip}

\begin{IEEEbiography}[{\includegraphics[width=1in,height=1.25in,clip,keepaspectratio]{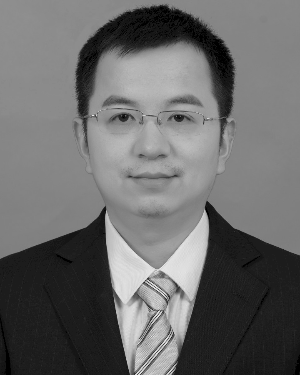}}]{Wenyong Yu}
received an M.S. degree and a Ph.D. degree from Huazhong University of Science and Technology, Wuhan, China, in 1999 and 2004, respectively. He is currently an Associate Professor with the School of Mechanical Science and Engineering, Huazhong University of Science and Technology.

His research interests include machine vision, intelligent control, and image processing.
\end{IEEEbiography}
\vspace{1\baselineskip}

\end{document}